\documentclass[journal]{IEEEtran}
\pdfoutput=1

\usepackage{amsmath,amssymb} % define this before the line numbering.
\usepackage{color}
\usepackage{enumitem}
\usepackage{algorithm} %ctan.org\pkg\algorithms
\usepackage{algpseudocode}
\usepackage{multirow}
\usepackage{footnote}
\usepackage{wrapfig}
\usepackage[table,xcdraw]{xcolor}
\usepackage{graphicx}
\usepackage{subcaption}
\usepackage[table,xcdraw]{xcolor}
\usepackage{hyperref}
\usepackage{multicol}
\usepackage{ntheorem}
\theoremseparator{.}

\newtheorem{observation}{Observation}
\ifCLASSINFOpdf
\else
\fi
\definecolor{ceruleanblue}{rgb}{0.16, 0.32, 0.75}
\newtheorem{definition}{Definition}

\newcommand\crule[3][black]{\textcolor{#1}{\rule{#2}{#3}}}
\newcommand\T{\rule{0pt}{2.6ex}}       % Top strut
\newcommand\B{\rule[-1.2ex]{0pt}{0pt}} % Bottom strut

\begin{document}
%
% paper title
% Titles are generally capitalized except for words such as a, an, and, as,
% at, but, by, for, in, nor, of, on, or, the, to and up, which are usually
% not capitalized unless they are the first or last word of the title.
% Linebreaks \\ can be used within to get better formatting as desired.
% Do not put math or special symbols in the title.
\title{Mirror, Mirror, on the Wall, Who's Got the Clearest Image of Them All? $-$ A Tailored Approach to Single Image Reflection Removal}
%
%
% author names and IEEE memberships
% note positions of commas and nonbreaking spaces ( ~ ) LaTeX will not break
% a structure at a ~ so this keeps an author's name from being broken across
% two lines.
% use \thanks{} to gain access to the first footnote area
% a separate \thanks must be used for each paragraph as LaTeX2e's \thanks
% was not built to handle multiple paragraphs
%

\author{Daniel~Heydecker$^*$, Georg~Maierhofer$^*$, Angelica~I.~Aviles-Rivero$^*$\thanks{$^*$These three authors contributed equally and hold joint first authorship. }, \\ Qingnan~Fan, Dongdong Chen, Carola-Bibiane~Sch{\"o}nlieb and  Sabine S{\"u}sstrunk
\thanks{Daniel~Heydecker, Georg~Maierhofer, Angelica~I.~Aviles-Rivero and Carola-Bibiane~Sch{\"o}nlieb are with the DAMTP and DPMMS, University of Cambridge. \{dh489,gam37,ai323,cbs31\}@cam.ac.uk}% I think Georg and I need to cite the grant number from EPSRC.
\thanks{Qingnan~Fan is with the Computer Science and Technology School, Shandong University. fqnchina@gmail.com; Dongdong Chen is with the University of Science and Technology of China. cddlyf@gmail.com}%
\thanks{Sabine~S{\"u}sstrunk is with the School of Computer and Communication Sciences, \'{E}cole Polytechnique F\'{e}d\'{e}rale de Lausanne. sabine.susstrunk@epfl.ch }
%\thanks{Manuscript received April 19, 2005; revised August 26, 2015.}}
}
% \author{
% \institute{$^{1}$University of Cambridge; $^{2}$Shandong University; $^{3}$EPFL\\
% \{dh489,gam37,ai323,cbs31\}@cam.ac.uk; fqnchina@gmail.com; sabine.susstrunk@epfl.ch}

% note the % following the last \IEEEmembership and also \thanks -
% these prevent an unwanted space from occurring between the last author name
% and the end of the author line. i.e., if you had this:
%
% \author{....lastname \thanks{...} \thanks{...} }
%                     ^------------^------------^----Do not want these spaces!
%
% a space would be appended to the last name and could cause every name on that
% line to be shifted left slightly. This is one of those "LaTeX things". For
% instance, "\textbf{A} \textbf{B}" will typeset as "A B" not "AB". To get
% "AB" then you have to do: "\textbf{A}\textbf{B}"
% \thanks is no different in this regard, so shield the last } of each \thanks
% that ends a line with a % and do not let a space in before the next \thanks.
% Spaces after \IEEEmembership other than the last one are OK (and needed) as
% you are supposed to have spaces between the names. For what it is worth,
% this is a minor point as most people would not even notice if the said evil
% space somehow managed to creep in.

% The paper headers
\markboth{Journal of \LaTeX\ Class Files,~Vol.~14, No.~8, August~2018}%
{Shell \MakeLowercase{\textit{et al.}}: Bare Demo of IEEEtran.cls for IEEE Journals}
% The only time the second header will appear is for the odd numbered pages
% after the title page when using the twoside option.
%
% *** Note that you probably will NOT want to include the author's ***
% *** name in the headers of peer review papers.                   ***
% You can use \ifCLASSOPTIONpeerreview for conditional compilation here if
% you desire.

% If you want to put a publisher's ID mark on the page you can do it like
% this:
%\IEEEpubid{0000--0000/00\$00.00~\copyright~2015 IEEE}
% Remember, if you use this you must call \IEEEpubidadjcol in the second
% column for its text to clear the IEEEpubid mark.

% use for special paper notices
%\IEEEspecialpapernotice{(Invited Paper)}

% make the title area
\maketitle

% As a general rule, do not put math, special symbols or citations
% in the abstract or keywords.
\begin{abstract}
Removing reflection artefacts from a single image is a problem of both theoretical and practical interest, which still presents challenges because of the massively ill-posed nature of the problem. In this work, we propose a technique based on a novel optimisation problem. Firstly, we introduce a \emph{simple} user interaction scheme, which helps minimise information loss in reflection-free regions. Secondly, we introduce an $H^2$ fidelity term, which preserves fine detail while enforcing global colour similarity.  We show that this combination allows us to mitigate some major drawbacks of the existing methods for reflection removal. We demonstrate, through numerical and visual experiments, that our method is able to outperform the state-of-the-art methods and compete with recent deep-learning approaches.
\end{abstract}

% Note that keywords are not normally used for peerreview papers.

% For peer review papers, you can put extra information on the cover
% page as needed:
% \ifCLASSOPTIONpeerreview
% \begin{center} \bfseries EDICS Category: 3-BBND \end{center}
% \fi
%
% For peerreview papers, this IEEEtran command inserts a page break and
% creates the second title. It will be ignored for other modes.
\IEEEpeerreviewmaketitle

\section{Introduction}
\begin{figure*}[t]
\centering
\includegraphics[width=1\textwidth]{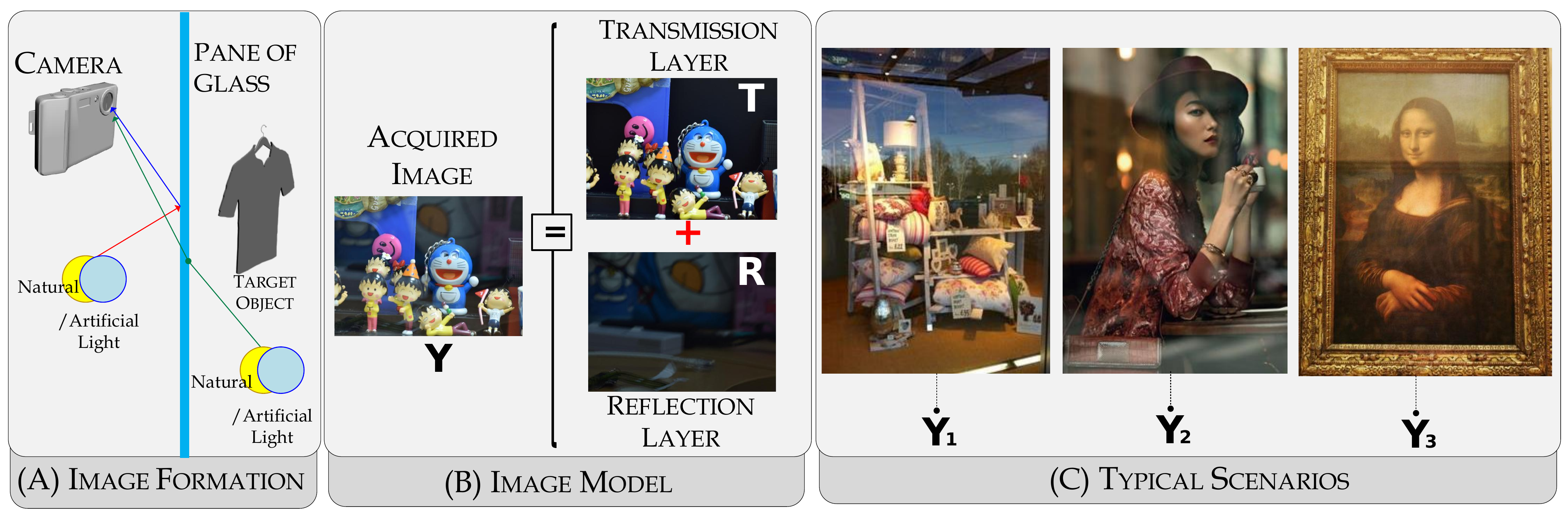}
\caption{(A) An illustration of the image formation in which  a target object captured through a pane of glass will have reflection artefacts. (B) Based on the image model, an acquired image ($\mathbf{Y}$) can be decomposed into two layers: Transmission ($\mathbf{T}$) and Reflection ($\mathbf{R}$).  (C) images ($\mathbf{Y}_{1,2,3}$) show a set of typical situations where there is no option but to take the picture through a pane of glass such as store display or in museums. %\textbf{\textcolor{red}{We may need to cut a-b to save space; don't think they add much}}
}
\label{fig::initialScheme}
\end{figure*}

This paper addresses the problem of single image reflection removal. Reflection artefacts are ubiquitous in many classes of images; in real-world scenes, the conditions are often far from optimal, and photographs have to be taken in which target objects are covered by reflections and artefacts appear in undesired places. This does not only affect amateur photography; such artefacts may also arise in documentation in museums and aquariums, or black-box cameras in cars (see Fig.~\ref{fig::initialScheme}). It is therefore unsurprising that the problem of how to remove reflection artefacts is of great interest, from both practical and theoretical points of view.

Although it is possible to reduce reflection artefacts by the use of specialised hardware such as polarisation filters ~\cite{schechner2000separation,agrawal2005removing,kong2014physically}, this option has several downsides. Firstly, even though the use of hardware can have a significant effect on removing the reflection, it only works when certain capture conditions are fulfilled, such as Brewster's angle \cite{lakhtakia1992general}. In practise, it is difficult to achieve optimal capture conditions, which results in residual reflections~\cite{farid1999separating,kong2011high}. As a result, post-processing techniques are often needed for further improvement of the image. Moreover, for the purposes of amateur photography, the use of specialised hardware is expensive, and consequently less appealing.

As an alternative to the use of specialised hardware, a body of research has established a variety of computational techniques. These can be divided in those that use \emph{multiple images}, and those that use a  \emph{single image}. The former techniques employ images from various view points (e.g.~\cite{szeliski2000layer,sarel2004separating,sinha2012image,guo2014robust}), with the aim of exploiting temporal information to separate the reflection artefacts from the observed target, while for the latter, carefully selected image priors are used to obtain a good approximation of the target object, for example~\cite{levin2007user,li2014single,wan2016depth,arvanitopoulos2017single}.

Although the use of multiple images somewhat mitigates the massively ill-posed problem created by the reflection removal formulation, the success of these techniques requires multiple images from several viewpoints and their performance is strongly conditional on the quality of the acquired temporal information. Moreover, in practice, acquisition conditions are non-optimal, which often results in image degradation, causing occlusions and blurring in the images. Therefore, either many images or post-processing are needed, which strongly restricts the applicability and feasibility of these methods to a typical end-user.
These constraints make single-image methods a focus of great attention to the scientific community, since it is appropriate for most users, and this is the approach which we will take in this paper.

Mathematically, an image $\mathbf{Y}$ containing reflection artefacts can be represented as a linear superposition~\cite{barrow1978recovering} as:

\begin{equation}
\label{eq:basicmodelfornoisyimage}
\mathbf{Y}=\mathbf{T}+ \mathbf{R},
\end{equation}

\noindent
where $\mathbf{T, R}$ are $n\times m$ matrices representing the transmission layer and reflection layer, respectively. Therefore, the goal of a reflection suppression technique is to approximate $\mathbf{T}$ from the acquired image $\mathbf{Y}$.

Although the body of literature for single-image reflection removal has proven promising results, this remains an open problem, and there is still potential for further enhancements. We consider the problem of how to get a better approximation of $\mathbf{T}$.

In this work, we propose a new approach, closely related to ~\cite{arvanitopoulos2017single}, and inspired by the observation that even \emph{low-level} user input may contain a lot of information. Our technique relies on additional information, which gives the rough location of reflections. In our experiments, this is given by user-input; in principle, this could be done by an algorithmic or machine-learning technique. We recast the reflection removal problem as an optimisation problem which is solved  iteratively, by breaking it up into two more computationally tractable problems. Compared to existing solutions from the literature, we achieve a better approximation of $\mathbf{T}$ from a well-chosen optimisation problem, while simultaneously preserving image details and eliminating global colour shifts. Our contributions are as follows:

\begin{itemize}
	\item  We propose a computationally tractable mathematical model for single-image reflection removal, in which we highlight:
      \begin{itemize}

      	\item A \emph{simple and tractable} user interaction method to select reflection-heavy regions, which is implemented at the level of the optimisation problem as a spatially aware prior term. We show that this improves the retention of detail in reflection-free areas.

             	 \item A combined \emph{$H^2$ fidelity term}, which combines $L^2$ and Laplacian terms. We show that this combination yields significant improvements in the quality of the colour and structure preservation.
      \end{itemize} We establish that the resulting optimisation problem can be solved efficiently by half-quadratic splitting.
  \item We validate the theory with a range of numerical and visual results, in different scenes and under varying capture conditions.
  \item We demonstrate that the combination of our fidelity term and prior term leads to a better approximation of $\mathbf{T}$ than state-of-the-art model based techniques, and can compete with the most recent deep-learning (DL) techniques.
\end{itemize}

\section{Related Work}
\label{subsec:background}

The problem of image reflection removal has been extensively investigated in the computer vision community, in which solutions rely on using multiple images and  single image data, alone or in combination with specialised hardware.
In this section, we review the existing techniques in turn.

\medskip

A number of techniques have been developed which use information from multiple images to detect and remove reflections. These include the use of different polarisation angles \cite{farid1999separating,schechner2000polarization,kong2011high,kong2012physically,kong2014physically}, adjustment of focal and flash settings \cite{schechner2000blind,schechner2000separation,agrawal2005removing}, and the uses of relative motion and coherence \cite{szeliski2000layer,sarel2004separating,li2013exploiting,sun2016automatic,xue2015computational,nandoriya2017video,shah2017removal,simon2015reflection,cheong2017reflection}. A recent technique \cite{han2018glass} seeks to improve on these methods by seeking to match the \emph{transmitted} layer, while other techniques may erroneously match the reflected layer.  Each of these techniques requires particular modelling hypotheses to be met, and advantageous capture conditions which may not be feasible in practice. \iffalse In contrast, the simple image capture settings requirements for taking a single-image, from a conventional camera, have motivated the fast development of techniques for the reflection removal problem. \fi

\medskip
 We now review the related works in single image techniques, as they are most applicable to everyday capture. A commonality of these techniques is the choice of a sparse gradient prior, which imposes a preference for output transmission layers, $\mathbf{T}$, with few strong edges.

A user-intervention method was proposed in~\cite{levin2007user}, which labels gradients as belonging to either transmission or reflection layer. They then propose to solve a constrained optimisation problem, with prior distribution given by the superposition of two Laplace distributions. A similar optimisation problem is used by~\cite{wan2016depth}, which replaces user-intervention labelling by a depth-of-field based inference scheme, while~\cite{shih2015reflection} relies on ghosting artefacts.

Our work is most closely related to the optimisation-based models and techniques of ~\cite{li2014single,arvanitopoulos2017single}. The authors of~\cite{li2014single} propose a smooth gradient prior on the reflection layer, and a sparse gradient prior on the transmission layer. This approach was adapted by Arvanitopoulos et al. in~\cite{arvanitopoulos2017single}, who proposed a Laplacian-based fidelity term with a novel sparse gradient prior. This preserves (Gestalt) continuity of structure, while also reducing loss of high-frequency detail in the transmission layer. The algorithm they propose is both more effective, and more computationally efficient, than the other techniques discussed above.
%% - deep learning
%deep learning (DL) is a recently emerging technique which has led to breakthroughs in image processing and related areas ~\cite{lecun2015deep}.
%\medskip \\
%\textbf{Machine-Learning Reflection Removal.}

The application of deep learning  to reflection removal was pioneered by Fan et al. in \cite{fangeneric}. In this work, the authors propose a deep neural network structure, which firstly predicts the edge map and then separates the layers. This technique outperforms the algorithmic approach of~\cite{li2014single}.
Further work in this direction was made by Zhang et al. \cite{zhang2018single}, who use a fully convolutional neural network with three loss terms, which help to ensure preservation of features and pixel-wise separation of the layers. Wan et al. \cite{wan2018crrn} seek to use a loss function inspired by human perception to estimate the gradient of the transmission layer, and use this to concurrently estimate the two layers using convolutional neural networks, and Jin et al. \cite{jin2018learning} proposes a convolutional neural network with a resampling strategy, to capture features of global priors, and avoid the ambiguity of the average colour. Most recently, Yang et al \cite{yang2018seeing} propose a bidirectional deep learning-scheme based on a cascade neutral network. This method first estimates the background layer $\mathbf{T}$, then uses this to estimate the reflected layer $\mathbf{R}$. Finally, the estimate on $\mathbf{R}$ is used to improve the estimate of $\mathbf{T}$.

%While in this paper the deep-learning based techniques provide an important benchmark, their classification as `single-image' techniques raises definitional issues. Deep-Learning schemes require a large set of training data, which raises problems of potential overfitting, and may be impractical for an end user. Moreover, at the time of writing, many of the deep-learning based approaches are more generic, and so may not be adapted to the specific problems of reflection removal, such as preserving detail and colour.

%Each of the works \cite{zhang2018single,wan2018crrn,jin2018learning,yang2018seeing} improves on the results of Fan et al. \cite{fangeneric} and on existing algorithmic approaches, such as Li and Brown \cite{li2014single}. These machine-learning techniques neither require multiple images, nor strong modelling assumptions on the nature of the reflection, and so may be more widely applicable than the algorithmic techniques described above.
%\medskip \\
The philosophy of our approach is similar to that of \cite{levin2007user}. Motivated by the principle that \emph{humans are good at distinguishing reflections}, both our work and \cite{levin2007user} seek to exploit further user input to assist an algorithmic technique. However, we emphasise that we are the first to propose a \emph{simple and tractable} user interaction scheme: in evaluating our user interaction scheme in Section IV/E3, we will see that our user interaction scheme requires very little effort from the user, and that our algorithm performs well with even very crude selection. By contrast, the algorithm of \cite{levin2007user} requires much more effort, and a much more detailed input.
%, is the inspiration for part of the model proposed by \cite{wan2018crrn}, although no user interaction is used in this case.
\section{Proposed Method} %%Tailored Reflection Suppression
%In this section, we describe our proposed solution for single-image reflection removal which we recast as an optimisation problem.
This section contains the three key parts of the proposed mathematical model: (i) the combined Laplacian and $L^2$ fidelity term, (ii) a \emph{spatially aware} prior term, given by user input, and (iii) the implementation using quadratic splitting for computational tractability.

Although the model for an image with reflection artefacts described in~\eqref{eq:basicmodelfornoisyimage} is widely-used, our solution adopts the observation of~\cite{schechner2000separation,li2014single,arvanitopoulos2017single} that the reflection layer is less in focus and often blurred, which we formalise as follows:
\begin{observation}
  In many cases, the reflected image will be blurred, and out of focus. This may be the case, for instance, if the reflected image is at a different focal distance from the transmitted layer. Moreover, reflections are often less intense than the transmitted layer.
\end{observation}

\noindent
Based on this observation, the image model~\cite{schechner2000separation,li2014single} which we adapt is
\begin{equation} \label{eq: model of noisy image}
\mathbf{Y}=w \mathbf{T} + (1-w)(\mathbf{k}\star \mathbf{R}),
\end{equation}
\noindent
where $\star$ denotes convolution, $w$ is a weight $w \in [0,1]$ that controls the relative strength of reflections, and $\mathbf{k}$ is a blurring kernel.

%\medskip
%\noindent\textbf{3.1 Fidelity and Prior Terms.}
\subsection{Fidelity and Prior Terms.}
We begin by discussing the prior term. Loss of some detail, in reflection heavy regions, is to be expected, and is a result of the ill-posed nature of reflection suppression. We seek to use low-level user input to reduce the loss of detail \emph{in reflection-free regions}, motivated by the following observation:

\begin{observation}
   In many instances, the reflections are only present in a region of the image, and it is easy for an end user to label these areas. In regions where reflections are not present, all edges in $\mathbf{Y}$ arise from $\mathbf{T}$, and so should not be penalised in a sparsity prior. Moreover, in certain instances, it may be particularly important to preserve fine detail in certain regions.
\end{observation}
\noindent
For instance, for photographs containing a window, the reflections will only occur in the window, and not elsewhere in the image. To this end, we propose to incorporate a \emph{region selection function} $\boldsymbol{\phi}$, taking values in $[0,1]$, into a \emph{spatially aware prior:}
\begin{equation}\label{eq:priorTerm}
P(\boldsymbol{\phi}, \mathbf{T})=\sum_{i,j} \phi_{ij}1[\nabla_xT_{ij}\neq 0 \text{ or } \nabla_yT_{ij}\neq 0].
\end{equation}
Here, $1[..]$ denotes the indicator function for the set of indexes $(i,j)$ where one of the gradients $\nabla_x \mathbf{T}, \nabla_y \mathbf{T}$ is nonzero. We assume that the region selection function $\boldsymbol{\phi}$ is given by the user, along with the input. Although this is philosophically similar to the user intervention method of ~\cite{levin2007user}, our approach is drastically less effort-intensive: rather than labelling many edges, it is sufficient to (crudely) indicate which regions contain reflections. The practicalities of our technique will be discussed in Subsection C below. We will show that, by choosing $\phi_{ij} \approx 1$ on reflection-heavy regions and $\phi_{ij}\approx 0$ elsewhere, we can minimise the loss of detail in reflection-free areas, and avoid the `flattening' effect described above. We also note that a na\"ive attempt to apply the approach of \cite{arvanitopoulos2017single} to a region of the image produces noticeable colour shifts at the boundary of the selected region, which our spatially aware prior term avoids.
\medskip\\ We now consider the fidelity term, seeking to build on the Laplacian fidelity term proposed by~\cite{arvanitopoulos2017single}; this choice of fidelity term penalises over-smoothing, and enforces consistency in fine details. Although this improves on the $L^2$ fidelity term of Xu et al.~\cite{xu2011image}, one can still observe significant `flattening' effects. These arise when there is significant colour variation over a large area: individual gradients are weak, and are neglected by the technique of~\cite{arvanitopoulos2017single}. This results in the whole region being given the same value in the output $\mathbf{T}$, producing unrealistic and visually unappealing results. Moreover, we also note that  for any constant matrix $\mathbf{C}$ the Laplacian is invariant under the transformation $\mathbf{T} \mapsto \mathbf{T+C}$.
As a result, the algorithm proposed by~\cite{arvanitopoulos2017single} risks producing global colour shifts; at the level of the optimisation problem, this reflects the non-uniqueness of minimisers. To eliminate this possibility, we propose a combined fidelity term:

\begin{equation}\label{eq:dataTerm}
d_\gamma(\mathbf{T}, \mathbf{Y})= \left\|\Delta\mathbf{T}-\Delta\mathbf{Y}\right\|_2^2 +\gamma \|\mathbf{T}-\mathbf{Y}\|_2^2,
\end{equation}
where $\Delta\mathbf{T}$ is the discrete Laplacian defined as $\Delta\mathbf{T}=\nabla_{xx}\mathbf{T}+\nabla_{yy}\mathbf{T}$, and $\gamma$ is a positive parameter controlling the relative importance of the two terms. We will see, in numerical experiments, that this leads to results with more natural, saturated colours, and which are consequently more visually pleasing. We remark that other kernel filters are possible which would play the same role of measuring structure, such as the discrete gradient $\nabla$, or more complicated elliptic second-order operators; we use the Laplacian for the following reasons. Firstly, the Laplacian penalises loss of high-frequency detail more strongly than first order operators such as $\nabla$, as can be seen by moving to Fourier space, and so our choice will preserve high-frequency details well. Secondly, the Laplacian is a simple measure of structure, and which is invariant under the (natural) symmetry of rotation. \medskip \\
Combining the prior and fidelity terms, as defined in \eqref{eq:priorTerm} and \eqref{eq:dataTerm}, our optimisation problem is therefore
\begin{equation} \label{eq:optimisationProblem}
\mathbf{T}^*=\text{argmin}_\mathbf{T}\left\{\left\|\Delta\mathbf{T}-\Delta\mathbf{Y}\right\|_2^2+ \gamma \|\mathbf{T}-\mathbf{Y}\|_2^2+ \lambda P(\boldsymbol{\phi}, \mathbf{T})\right\}.
\end{equation}

Here, $\lambda$ is a regularisation parameter to be chosen later. The reader is invited to compare this optimisation problem to the similar problem of (localised) $L^0$ image smoothing, but to note the important difference of having a fidelity term including the image Laplacian. In the next section, we will detail how the proposed optimisation problem can be solved in a tractable computational manner by using quadratic splitting.

\iffalse Here, $\lambda$ is a regularisation parameter to be chosen later. The reader is invited to compare this optimisation problem to the equivalent problem for (localised) $L^0$ image smoothing. In the next section, we will detail how the proposed optimisation problem can be solved in a tractable computational manner by using quadratic splitting. \fi

%\medskip
%\noindent\textbf{3.2 Solving the Optimisation Problem.}
\subsection{Solving the Optimisation Problem.}
We solve the optimisation problem introduced in~(\ref{eq:optimisationProblem}) by half-quadratic splitting. We introduce auxiliary variables $\mathbf{D}^x,\mathbf{D}^y$ as proxies for, respectively, $\nabla_x \mathbf{T}$ and $\nabla_y \mathbf{T}$. For ease of notation, we write $\mathbf{D}$ for the pair $[\mathbf{D}^x,\mathbf{D}^y]$, and similarly $\nabla \mathbf{T}$ for the pair $[\nabla_x\mathbf{T},\nabla_y\mathbf{T}]$. This leads to the auxiliary problem:
\begin{equation}
\begin{aligned} \label{eq: aux problem}
\mathbf{T^*, D^*}=&\text{argmin}_\mathbf{T,D}\left\{\left\|\Delta\mathbf{T}-\Delta\mathbf{Y}\right\|_2^2+ \gamma \|\mathbf{T}-\mathbf{Y}\|_2^2 \right.\\
& \left.+\lambda P(\boldsymbol{\phi}, \mathbf{D})+\beta\|\mathbf{D}-\nabla \mathbf{T}\|_2^2\right\}
\end{aligned}
\end{equation}
where $\beta\in\mathbb{R}_{>0}$ is a penalty parameter yet to be chosen, and we use the shorthand
\begin{equation}
 P(\boldsymbol{\phi}, \mathbf{D}) = \sum_{i,j} \phi_{ij}1[D^x_{ij}\neq 0 \text{ or } D^y_{ij} \neq 0].
\end{equation}

\noindent
Notice that in the limit $\beta\rightarrow\infty$ the axillary penalty term ensures that we recover the solution to the original optimisation problem \eqref{eq:optimisationProblem}. Hence, we may solve the optimisation problem (\ref{eq: aux problem}) by splitting into two more computational tractable problems. We alternate between optimising over $\mathbf{T}$ and $\mathbf{D}$, while keeping the other fixed; at the same time, we increment $\beta$ so that, after a large number of steps, $\mathbf{D}$ is a good approximation of $\nabla \mathbf{T}$. We give details on the solution of each sub-problem below, and the full solution is presented in Algorithm \ref{alg::ours}.

\medskip
\noindent
\textbf{$\blacktriangleright$Sub-problem 1: Optimisation over $\mathbf{T}$.}
For a fixed $\mathbf{D}$, we wish to optimise:
\begin{equation} \label{eq: op over T}
\begin{aligned}
\mathbf{T}^*=&\text{argmin}_T\left\{\left\|\Delta\mathbf{T}-\Delta\mathbf{Y}\right\|_2^2+\gamma\|\mathbf{T}-\mathbf{Y}\|_2^2 \right.\\
&\left. +\beta\left\|\mathbf{D}-\nabla \mathbf{T}\right\|_2^2\right\}.
\end{aligned}
\end{equation}
\noindent
The objective function is now quadratic in $\mathbf{T}$. We note that the discrete gradient $\nabla$ and the discrete Laplacian $\Delta$ are both linear maps which take an $m\times n$ image matrix to an array of size $2\times m\times n$ and $m\times n$ respectively. We may thus view these linear maps as tensors, and use index notation to describe their action on an image $(T_{ij})$ as follows:

\begin{align}
(\nabla_\mu \mathbf{T})_{ij}=\sum_{k,l}\nabla^\mu _{ijkl}T_{kl}; \hspace{0.1cm} 1\leq i \leq m, \hspace{0.1cm}1\leq j\leq n, \hspace{0.1cm}\mu \in \{x,y\}
\end{align}
and similarly:
\begin{align}
(\Delta\mathbf{T})_{ij}=\sum_{k,l} \Delta_{ijkl}T_{kl}; \hspace{0.2cm} 1\leq i \leq m, \hspace{0.1cm} 1\leq j \leq n.
\end{align}

\begin{algorithm}[t!]
 \begin{algorithmic}[1]
% \SetKwInOut{Input}{input}\SetKwInOut{Output}{Output}
            \State Start from $\mathbf{T}\gets \mathbf{Y}$ and $\beta=\beta_\text{min}$\;
            \While{$\beta \leq \beta_\mathrm{max}$ }
            \State Optimise over $\mathbf{D}$, for the current value of $\mathbf{T}$:
                \begin{align*}
                \text{Set }(D^x_{ij},D^y_{ij})= \begin{cases} (0,0)  \text{ if \vspace{0.1cm}} |(\nabla_x {T}_{ij},\nabla_y {T}_{ij})|_2^2 \leq \frac{\lambda_{ij}}{\beta}; \\ (\nabla_x {T}_{ij},\nabla_y {T}_{ij})  \text{ o.w.}
                \end{cases}
            \end{align*}

            \State Using ADAM~\cite{kingma2014adam} and (\ref{eq: gradient component}), find the minimum $\mathbf{T^\star}$ of (\ref{eq: op over T}), and replace $\mathbf{T}\gets \mathbf{T}^\star$.\;

            \State Increment $\beta\gets \kappa \beta$\;

            \EndWhile
       	\State \textbf{return} $\mathbf{T}$.
 \end{algorithmic}
 \caption{Our Proposed Method}
 \label{alg::ours}
 \end{algorithm}

\noindent

With this notation, we can write the objective function as:
\begin{equation}
\begin{aligned}
&F_1(\mathbf{T}, \mathbf{D})=\beta \sum_{\substack{1\leq i\leq m \\  1\leq j\leq n \\  \mu \in \{x,y\}}} \left(D^\mu_{ij}-\sum_{1\leq k\leq m, 1\leq l \leq n} \nabla^\mu_{ijkl} T_{kl}\right)^2 + \\
&\sum_{i\leq m, j\leq n}\left(\left(\sum_{k\leq m, l\leq n} \Delta_{ijkl}(T_{kl}-Y_{kl})\right)^2+\gamma(T_{ij}-Y_{ij})^2 \right)
\end{aligned}
\end{equation}
We observe that this is quadratic, and in particular smooth, in the components $T_{ij}$. Using the summation convention, we compute the gradient:
\begin{multline} \label{eq: gradient component}
\frac{\partial}{\partial T_{ij}}F_1(\mathbf{T}, \mathbf{D})=  2\Delta_{abij}\Delta_{abkl}(T_{kl}-Y_{kl})+2\gamma(T_{ij}-Y_{ij})\\ +2\beta\nabla^\mu_{abij}(\nabla^\mu_{abkl}T_{kl}-D^\mu_{ab}).
\end{multline}
We use this computation, together with ADAM~\cite{kingma2014adam}, a first-order gradient descent method in stochastic optimisation, to efficiently optimise over $\mathbf{T}$.

\medskip
\noindent
\textbf{$\blacktriangleright$Sub-problem 2: Optimisation over $\mathbf{D}$.}
For a fixed $\mathbf{T}$, the optimisation problem in $\mathbf{D}$ is given by
\begin{equation} \label{eq: op over G}
\mathbf{D}^*=\text{argmin}_\mathbf{D}\left\{\beta \left\|\mathbf{D}-\nabla \mathbf{T}\right\|_2^2+ \lambda P(\boldsymbol{\phi},\mathbf{D})\right\}.
\end{equation}
Although the objective function, $F_2$, is neither convex nor smooth, due to the $L^0$ prior term, we observe that it separates as
\noindent
\begin{equation} \label{eq:separates}
\begin{aligned}
F_2(\mathbf{T}, \mathbf{D}) =& \sum_{i,j}  \left[\beta\left(|D^x_{ij}-\nabla_x T_{ij}|^2+|D^y_{ij}-\nabla_y T_{ij}|^2\right)\right.\\
&\left. +\lambda\phi_{ij} 1\left((D^x_{ij},D^y_{ij})\neq 0\right)\right].
\end{aligned}
\end{equation}
\noindent
%Hence, (\ref{eq: op over G}) separates as the sum of multiple 2-dimensional optimisation problems of the form:

%\begin{equation} \label{eq: toy problem}
%(x^*,y^*)=\text{argmin}_{(x,y)}\left\{|x-x_0|^2+|y-y_0|^2+\frac{\lambda}{\beta}1((x,y)\neq 0)\right\}.
%\end{equation}
%It is straightforward to verify that an explicit solution to (\ref{eq: toy problem}) is given by:
%\begin{equation}
%(x^*, y^*)=\begin{cases} (0,0) & \text{ if } |(x_0, y_0)|^2_2 \leq \frac{\lambda}{\beta}; \\ (x_0, y_0) & \text{ otherwise}.\end{cases} \end{equation}
By explicitly solving the separated problems for each pair $(D^x_{ij}, D^y_{ij})$, it is straightforward to see that \emph{a} solution to (\ref{eq:separates})  is given by
\begin{equation} \label{eq: formula for g} (D^x_{ij},D^y_{ij})= \begin{cases} (0,0) & \text{ if } |(\nabla_x {T}_{ij},\nabla_y {T}_{ij})|_2^2 \leq \frac{\lambda\phi_{ij}}{\beta}; \\ (\nabla_x {T}_{ij},\nabla_y {T}_{ij}) & \text{ otherwise}. \end{cases}  \end{equation}
Moreover, this minimiser is unique, provided that none of the edges are in the boundary case $|(\nabla_x {T}_{ij},\nabla_y {T}_{ij})|_2^2 = \frac{\lambda\phi_{ij}}{\beta}$.

\begin{figure}[t!]
\centering
\includegraphics[width=0.48\textwidth]{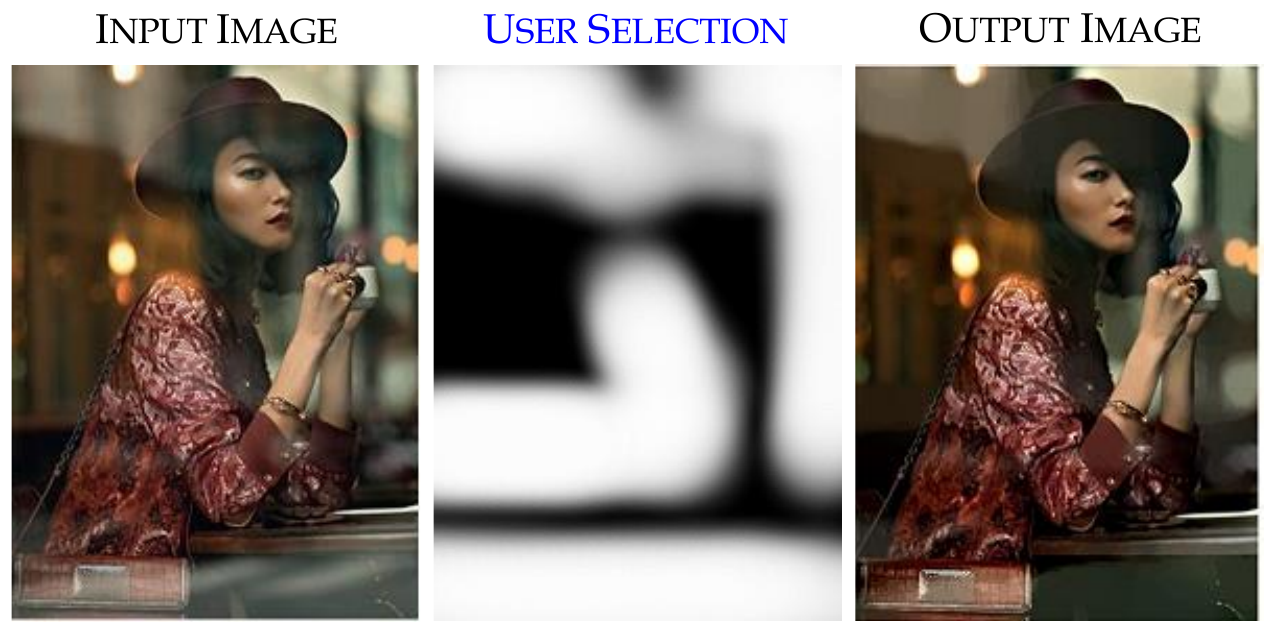}
\caption{From left to right. Input image, visualisation of the user interaction in practise and
output image with our technique.}
\label{fig::selRegions}
\end{figure}

\medskip Hence, the optimisation (\ref{eq: op over G}) removes gradients below the \emph{local} threshold $\frac{\lambda\phi_{ij}}{\beta}$. We will show, in numerical experiments, that this has the effect of smoothing \emph{only} the selected regions, while keeping the strong edges which force continuity of structures, as was described in Section \ref{subsec:background}.

The overall procedure of our method $-$ in which previous individual steps are combined to solve the original optimisation problem (\ref{eq:optimisationProblem}) $-$ is listed in Algorithm~\ref{alg::ours}.

\subsection{User Interaction Scheme.}
We describe the user interaction scheme, and how the region selection function $\phi_{ij}$ may be obtained in practice. We recall that $\mathbf{\phi}$ is responsible for passing information about the location of reflection into the algorithm, and that it takes values in  the range $[0,1]$ with
\begin{itemize}
\item $\phi_{ij}$ close to $1$ if a reflection is present at pixel $(i,j)$ and
\item $\phi_{ij}$ close to $0$ if no reflection is present at pixel $(i,j)$.
\end{itemize}
In practise a user, or an arbitrary instance that can recognise rough locations of reflections, is given an image, as in left-side of Fig. \ref{fig::selRegions}, and selects the regions in which reflections are present. A possible result can be seen in the middle part of Fig. \ref{fig::selRegions}, where the values of $\phi_{ij}$ are displayed as the grey-values in the image. This selection is then fed into our algorithm together with the input image to produce the reflection removed output as shown at right side of Fig.\ref{fig::selRegions}.

In the absence of user interaction, we default to $\phi_{ij}\equiv 1$; that is, we assume reflections are present throughout the image.

It is noteworthy that the way this selection is performed is very simple and requires little effort. This makes it suitable for a range of applications, from an amateur human user, to algorithms that can recognise reflections, even in a very crude manner. For our experiments, the selection was performed by creating an overlay image in a raster graphics editor, where white regions are marked with a rough brush on top of reflections. This process can be performed in a matter of seconds for each image. The results can, of course, improve with increasing selection quality, but even a rough selection produces significant improvements over no selection; see Section IV/E3 for experiments and discussion. Examples of region selection in practice are included in Section IV of the supplementary material.

% For instance in Fig. \ref{fig:ours_compare_selection} we consider two cases - OURS (W/O) where $\phi_{ij}=1$ uniformly and OURS (W) where the reflections have been more carefully marked. It is possible to observe a clear improvement in the quality of the outcome, both from a visual, as well as from a numerical point of view.

\subsection{Performance Reasoning of  Parameters}  \label{sec:proposed method D}
Our procedure uses two parameters $\lambda, \gamma$, and an auxiliary parameter $\beta$ in intermediary optimisation steps. We think of $\beta$ as a coupling parameter, which determines the importance of the texture term in comparison to the coupling to the auxiliary variable $\mathbf{D}$. At later iterations, $\beta$ is large and the coupling is strong, which justifies the use of $\mathbf{D}$ as a proxy for $\nabla \mathbf{T}$.

\begin{figure*}[t!]
\centering
\includegraphics[width=1\textwidth]{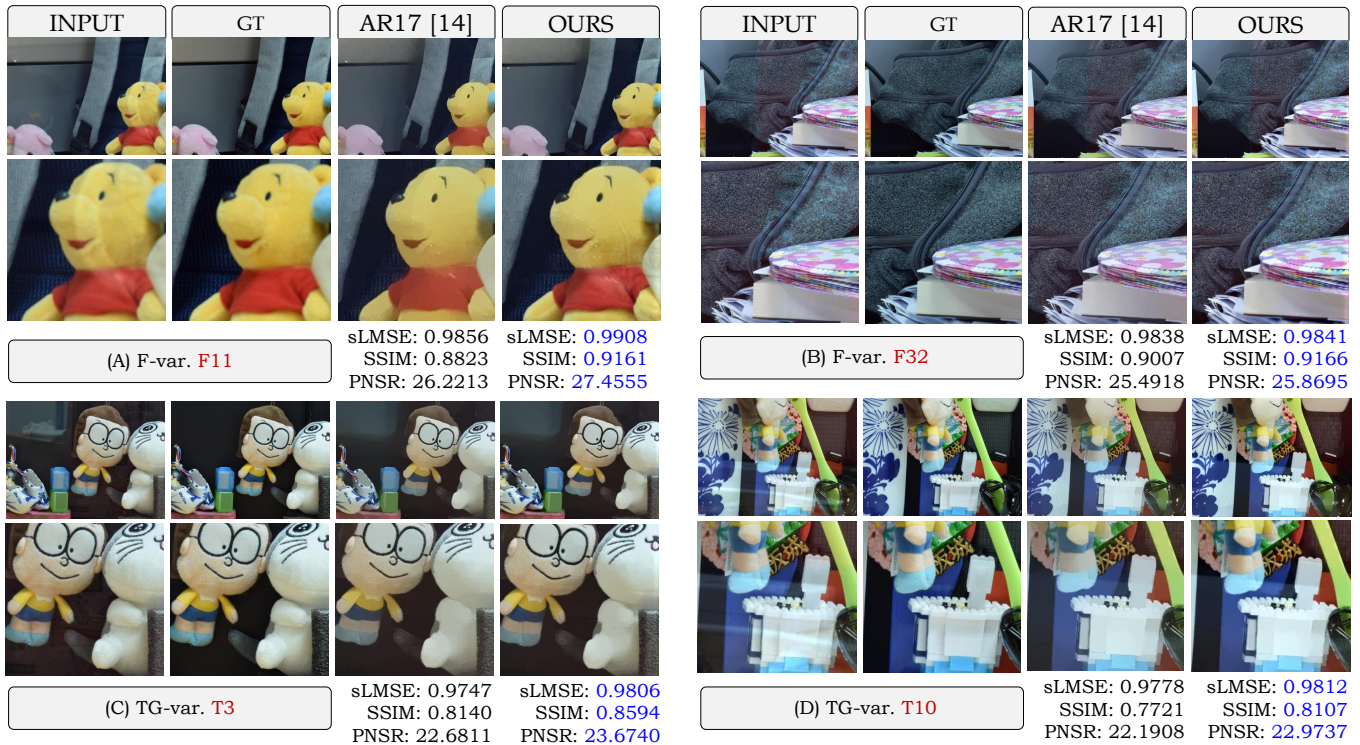}
\caption{(E1). Examples of the output, along with ground truth, of our approach compared against AR17~\cite{arvanitopoulos2017single}. The examples with varying settings such as the focus in (A) and (B) and the glass thickness in (C) and (D).  The three evaluation metrics of the reflection-free image are computed using the ground truth.}
\label{fig:OURSvsSIRS}
\end{figure*}

The parameter $\lambda$ determines the relative importance of preserving the structure versus preserving the texture. In terms of the model described above, it controls the importance of the penalty term $P(\boldsymbol{\phi},\mathbf{T})$ against the Laplacian $\|\Delta\mathbf{T}-\Delta\mathbf{Y}\|_2^2$. In regions where $\lambda\phi_{ij}$ is comparatively large, the sparsity of edges is much more important than the texture. Therefore, any edges which do not enforce structure will be washed out, and the region is smoothed during the optimisation over $\mathbf{D}$. On the other hand, in regions where $\lambda\phi_{ij}$ is comparatively small, the texture term dominates, and only very few edges are removed. In terms of the algorithm, this corresponds to controlling the edge threshold $\frac{\lambda\phi_{ij}}{\beta}$. This is illustrated in the supplementary material.

\medskip
We also give an interpretation of why it is natural to increase $\beta$ in this way. In the first stages of the iteration, $\beta$ is very small, and so the threshold keeps only the largest magnitude edges, and sets most edges of reflection-heavy areas to $0$. After each iteration, $\beta$ increases and the threshold $\frac{\lambda_{ij}}{\beta}$ decreases, and so the next iteration will preserve more edges. \emph{Hence, in reflection-heavy areas, we include edges in decreasing order of magnitude}; this corresponds to looking at strongly-defined structures first, and then considering incrementally weaker structure. This is illustrated in the supplementary material.

We give a theoretical basis for excluding the limiting regimes of either $\gamma \ll 1$ or $\gamma \geq 1$. In the regime where $\gamma \ll 1$, we may consider a step of the gradient descent to be a step of `uncorrected' gradient descent, with $\gamma=0$, followed by a small correction $\gamma(\mathbf{Y}-\mathbf{T})$ to correct colour shift. For this reason, if $\gamma\ll 1$ is too small, our algorithm will not adequately correct for colour shifts. On the other hand, if $\gamma>1$, then the $L^2$ term dominates the Laplacian term, and we expect blurring and loss of texture, as discussed in \cite{arvanitopoulos2017single}.

\section{Experimental Results}\label{sec:experiments}
In this section, we describe in detail the range of experiments that we conducted to validate our proposed method.

\subsection{Data Description.}
We evaluate the theory using the following three datasets.
Firstly, we use real-world data from the SIR$^2$ benchmark dataset~\cite{wan2017benchmarking}. The dataset is composed of 1500 images with size of  400$\times$540, and provides variety in scenes with different degrees of freedom in terms of aperture size and thickness of the glass. These variations allow us to test the respective algorithms in the presence of different effects, such as reflection shift. Moreover, it provides a ground truth that permits for quantitative evaluation. We also use the Berkeley dataset from ~\cite{zhang2018single}, which contains 110 real image pairs (reflection and transmission layer) whose characteristics can be founds in~\cite{zhang2018single}. Finally, we also use a selection of `real-world' images from \cite{fangeneric}, for which ground truths are not available.
All measurements and reconstructions were taken from these datasets.

\subsection{Evaluation Methodology.} \label{sec:evaluation_methodology}
We design a four-part evaluation scheme, where the  evaluation protocol for each part is as follows. \\
\textbf{(E1)} The first part is a visual comparison of our method against AR17~\cite{arvanitopoulos2017single}. We remark that in the case $\gamma=0, \phi=1$, our method reduces to that of AR17; this comparison therefore shows that the changes made to the objective function fulfil their intended purposes. \\
\textbf{(E2)} The main part of the evaluation is to compare our solution to the state-of-the-art methods. In (E2a) we compare to state-of-the-art algorithmic techniques LB14~\cite{li2014single}, SH15~\cite{shih2015reflection}, AR17~\cite{arvanitopoulos2017single}, using FAN17~\cite{fangeneric} as a benchmark. (E2b) is an evaluation against more recent advances in deep-learning  FAN17~\cite{fangeneric}, WAN18~\cite{wan2018crrn}, ZHANG18~\cite{zhang2018single} and YANG18~\cite{yang2018seeing} on both real-world images and the Berkeley dataset. We present both numerical comparisons, averaged over the SIR$^2$ and Berkely datasets in (E2a, E2b) respectively, and visual comparisons for a range of selected images from all three datasets.\\
\textbf{(E3)}  We evaluate the impact of the user input, and show the results of our method with no region selection, with crude region selection and with more detailed region selection. This will justify our claim that crude region selection is sufficient to minimise loss of detail in reflection-free areas, but offers a substantial qualitative improvement on \emph{no} region selection. \\
\textbf{(E4)} Finally, we demonstrate that, by comparison to the existing user interaction approach of Levin\cite{levin2007user}, we produce better results whilst requiring less effort from the end-user.

%\iffalse \textbf{(E5)} Finally, we address the limitations of our proposed method and discuss the behaviour of different techniques in failure cases. \fi

We address our scheme from both qualitative and quantitative points of view. The former is based on a visual inspection of the output $\mathbf{T}$,  and the latter on the computation of three metrics: the structural similarity (SSIM) index~\cite{wang2004image}, the Peak Signal-to-Noise Ratio (PSNR) and the inverted Localised Mean Squared Error (sLMSE). Explicit definition of the metrics can be found in Section VI of the Supplemental Material.

\begin{figure*}[t!]
\centering
\includegraphics[width=0.98\textwidth]{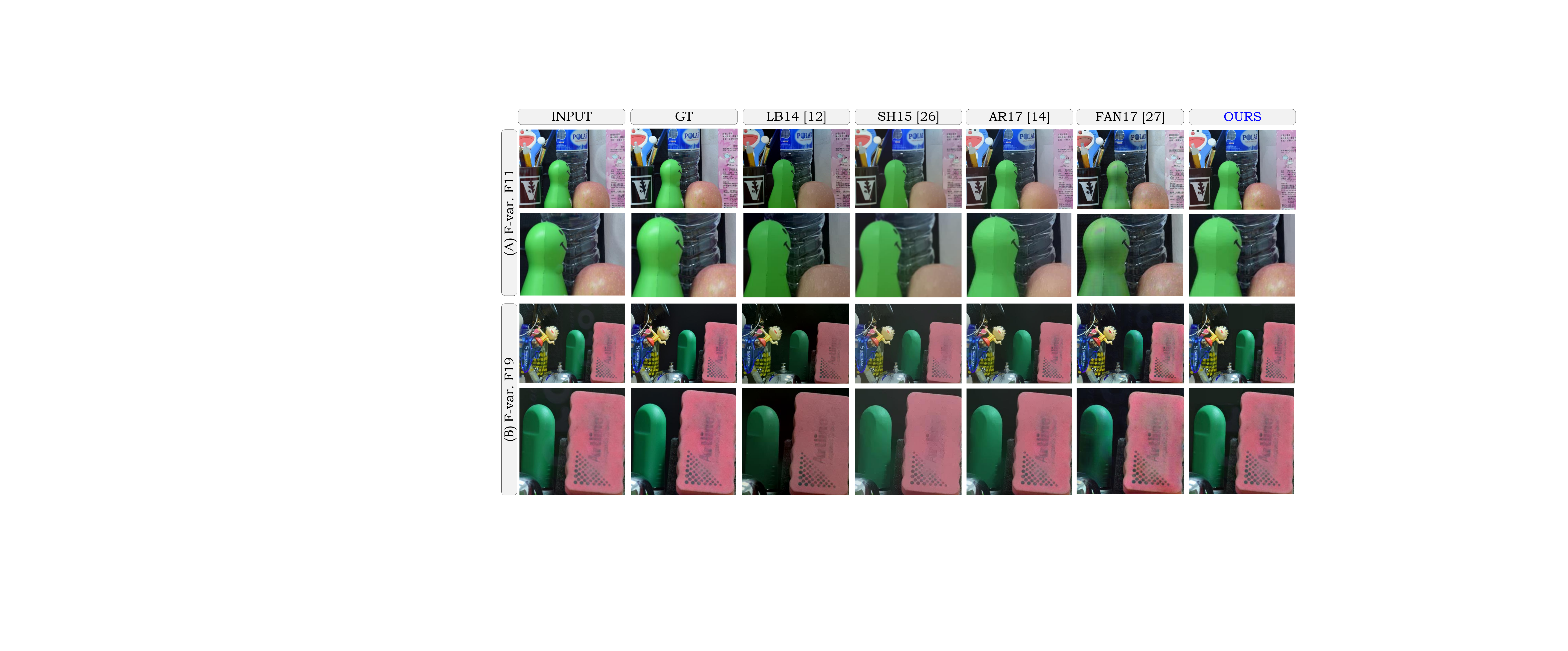}
\caption{(E2a). Visual comparison against the state-of-the-art of model-based approaches (including FAN17~\cite{fangeneric} as baseline for comparison). The selected frames show variations in shape, colour and texture to appreciate the performance of the compared approaches. Overall, our approach gives a better approximation of $\mathbf{T}$ by preserving colour and structure quality while keeping fine details. Details are better appreciated on screen.}
\label{fig::oursVSstateoftheart}
\end{figure*}

\begin{table*}[t!]
\centering
\resizebox{0.85\textwidth}{!} {
\begin{tabular}{|c|c|c|c|c|c|c|c|c|c|}
\hline
\cellcolor[HTML]{EFEFEF}{\color[HTML]{000000} } & \multicolumn{3}{c|
}{\cellcolor[HTML]{EFEFEF}{\color[HTML]{000000} sLMSE}} & \multicolumn{3}{c|}{\cellcolor[HTML]{EFEFEF}{\color[HTML]{000000} SSIM}} & \multicolumn{3}{c|}{\cellcolor[HTML]{EFEFEF}PNSR} \\ \cline{2-10}
\multirow{-2}{*}{\cellcolor[HTML]{EFEFEF}{\color[HTML]{000000} \textbf{F-var.}}} & F11 & F19 & F32 & F11 & F19 & F32 & F11 & F19 & F32 \\ \hline\hline
LB14~\cite{li2014single} & 0.835  & 0.832 & 0.833 & 0.784  & 0.804 & 0.791 & 21.659 & 21.869 & 21.678 \\
 SH15~\cite{shih2015reflection} & 0.901 & 0.852 & 0.874 & 0.779 & 0.813 & 0.765 & 21.642 & 22.046 & 21.620\\
AR17~\cite{arvanitopoulos2017single} & 0.983 & 0.984 & \cellcolor[HTML]{9AFF99}\textbf{0.984} & 0.820 & 0.825 & 0.824 & 22.748 & 22.705 & 22.851\\
FAN17~\cite{fangeneric} & 0.981 & 0.982 & 0.982 & \cellcolor[HTML]{9AFF99}\textbf{0.854} & 0.859 & 0.851 &\cellcolor[HTML]{9AFF99}\textbf{ 23.262} & 23.853 & 23.432\\
\textbf{OURS} & \cellcolor[HTML]{9AFF99} \textbf{0.984} & \cellcolor[HTML]{9AFF99}\textbf{0.986} & \cellcolor[HTML]{9AFF99}\textbf{0.984} & 0.852 & \cellcolor[HTML]{9AFF99}\textbf{0.866} & \cellcolor[HTML]{9AFF99}\textbf{0.854} & 23.254 & \cellcolor[HTML]{9AFF99}\textbf{23.907} & \cellcolor[HTML]{9AFF99}\textbf{23.649}\\\hline
\end{tabular}
} \\ \vspace{0.15cm}
\centering
\resizebox{0.85\textwidth}{!} {
\begin{tabular}{|c|c|c|c|c|c|c|c|c|c|}
\hline
\cellcolor[HTML]{EFEFEF}{\color[HTML]{000000} } & \multicolumn{3}{c|
}{\cellcolor[HTML]{EFEFEF}{\color[HTML]{000000} sLMSE}} & \multicolumn{3}{c|}{\cellcolor[HTML]{EFEFEF}{\color[HTML]{000000} SSIM}} & \multicolumn{3}{c|}{\cellcolor[HTML]{EFEFEF}PNSR} \\ \cline{2-10}
\multirow{-2}{*}{\cellcolor[HTML]{EFEFEF}{\color[HTML]{000000} \textbf{TG-var.}}} & TG3 & TG5 & TG10 & TG3 & TG5 & TG10 & TG3 & TG5 & TG10 \\ \hline\hline
LB14~\cite{li2014single}& 0.834 & 0.833 & 0.834 & 0.718 & 0.811 & 0.805 & 21.605 & 21.981 & 21.850 \\
 SH15~\cite{shih2015reflection} & 0.915 & 0.889 & 0.917 & 0.779 & 0.820 & 0.765 & 21.682 & 22.546 & 21.620  \\
AR17~\cite{arvanitopoulos2017single} & 0.983 & \cellcolor[HTML]{9AFF99}\textbf{0.984} & 0.982 & 0.820 & 0.825 & 0.824 & 22.748 & 22.705 &  22.851 \\
FAN17~\cite{fangeneric} & 0.981 & 0.981 & 0.981 & \cellcolor[HTML]{9AFF99}\textbf{0.850} & \cellcolor[HTML]{9AFF99}\textbf{0.852} & 0.852 & \cellcolor[HTML]{9AFF99}\textbf{23.415} & 23.403 & 23.470 \\
\textbf{OURS} & \cellcolor[HTML]{9AFF99}\textbf{0.984} & \cellcolor[HTML]{9AFF99}\textbf{0.984} & \cellcolor[HTML]{9AFF99}\textbf{0.984} & 0.846 & 0.851 & \cellcolor[HTML]{9AFF99}\textbf{0.861} & 23.374 & \cellcolor[HTML]{9AFF99}\textbf{23.421} & \cellcolor[HTML]{9AFF99}\textbf{23.507}\\\hline
\end{tabular}
}
\caption{(E2a). Measures averaged over all images in the solid-object dataset \cite{wan2017benchmarking}.}
\label{tab:measure-outcome}
\end{table*}

\begin{figure*}[h!]
\centering
\includegraphics[width=1\textwidth]{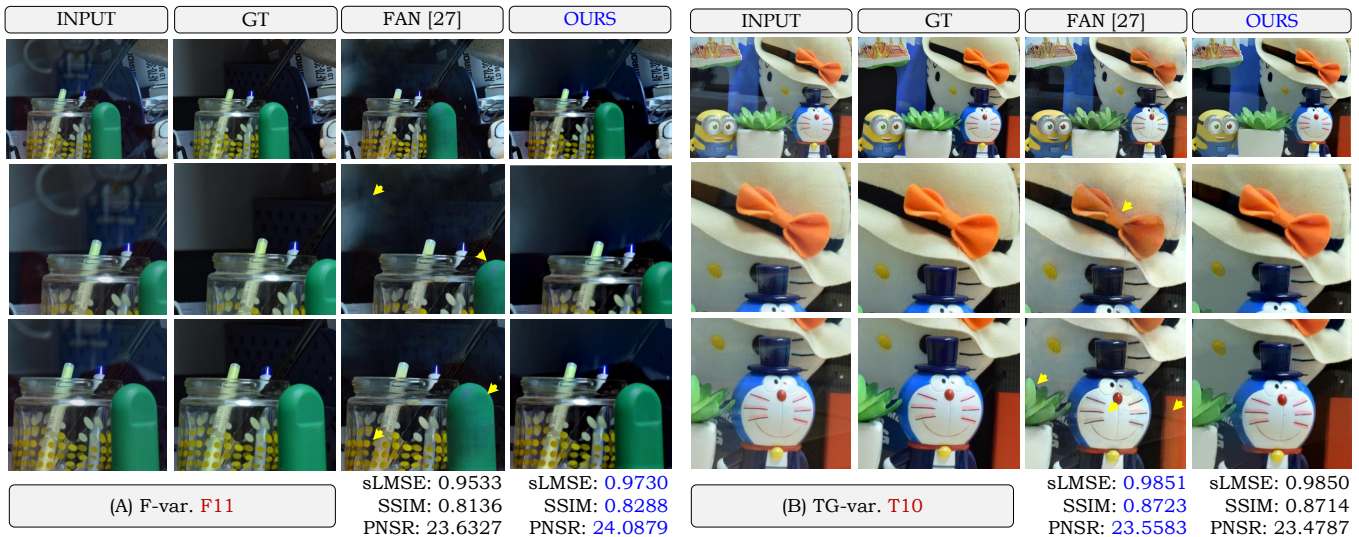}
\caption{(E2b). Two interesting cases in which we visually and numerically compare our approach against the work of Fan et al. \cite{fangeneric}. We emphasise that even in cases when the metrics are higher for FAN17~\cite{fangeneric}, the output from our algorithm appears visually more appealing and natural. We highlight the false colour effects (see bow in (B)), loss of fine details (see green object in (A)) and reflection artefacts (see yellow markers in both) in the output of FAN17. Details are better appreciated on screen.}
\label{fig::OURSvsFAN}
\end{figure*}

\subsection{Parameter Selection.}
For each of the approaches LB14~\cite{li2014single}, SH15~\cite{shih2015reflection} and AR17~\cite{arvanitopoulos2017single}, we set the parameters as described in the corresponding paper. Moreover, the comparison study was performed using the available codes from each corresponding author. For FAN17~\cite{fangeneric}, we assumed a given trained network and with parameters set as described in that paper.

For our approach, we set the values of the ADAM method as suggested in~\cite{kingma2014adam}. Moreover,  we set  $\lambda = 2e-3$, $\beta_\mathrm{max}=1e5$ and $\kappa=2$  and  $\gamma=0.012$. The choices of $\lambda, \beta_\mathrm{max}, \kappa$ follow \cite{arvanitopoulos2017single} for analogous parameters, which is consistent with the reasoning in Subsection \ref{sec:proposed method D}. $\gamma$ was chosen based on experimental results for a range of images disjoint from the test dataset, with a range of test values following the discussion in Subsection \ref{sec:proposed method D}

\subsection{Results and Discussion.} We evaluate our proposed method following the scheme described in Section \ref{sec:evaluation_methodology}.

\textbf{(E1).} We begin by evaluating our method against AR17~\cite{arvanitopoulos2017single}.  We ran both approaches on the complete solid objects category of the dataset.  In Fig.~\ref{fig:OURSvsSIRS}, we show four output examples with different settings (Aperture value F=$\{11,32\}$ and thickness of glass TG=$\{3,10\}$). Visual assessment agrees with the theory of our approach, in which we highlight the elimination of colour shifts and the preservation of the image details. Most notably, we see that our approach enforces global colour similarity and avoids blurring effects produced by the outputs of  AR17~\cite{arvanitopoulos2017single}; see, for example, outputs (A), (C) and (D). The detail in Fig.~\ref{fig:OURSvsSIRS} highlights these effects, in particular in (A) the blur and colour loss effects in the \emph{Winnie the Pooh} toy, in (C) the loss of edge details in the shirt collar (left toy) and the neck (white toy), and in (D) a blurring effect in the toy's legs.
In the detail of output (B), it can be seen that AR17~\cite{arvanitopoulos2017single} fails to preserve the shadows and the colours of the flowers. This is further reflected in the numerical results, where our method reported higher values for the three evaluation metrics.

Overall, we noticed that often AR17~\cite{arvanitopoulos2017single} fails to penalise colour shifts, due to the translation invariance of the Laplacian fidelity term. It also tends to produce blurring effects in reflection-free parts of the image, which our approach is able to prevent through our spatially aware technique.

\medskip
\textbf{(E2a).} We now evaluate our approach against the model-based state-of-the-art methods (LB14~\cite{li2014single}, SH15~\cite{shih2015reflection}, AR17~\cite{arvanitopoulos2017single}, and include FAN17~\cite{fangeneric} as a baseline of comparison) using the full solid objects category of the SIR$^2$ dataset. As discussed above, we may view the results of AR17 as those of our algorithm in the special case $\gamma=0$, and without user interaction $(\phi \equiv 1)$ to evaluate the effect of these changes. We emphasise that results for our algorithm were generated with user interaction, \emph{as this is a key part of our technique.}

We show the output of the selected methods and our proposed one for four chosen images along with the ground truth in Fig.~\ref{fig::oursVSstateoftheart}. By visual inspection, we observe that outputs generated with LB14~\cite{li2014single} are darker than the desired output; see, for instance, the detail of (A). Moreover, LB14  fails to preserve texture and global colour similarity,  as is apparent in (A) on the surface of the apple, and (B) on the pink block. By contrast, our approach was able to keep the details on both cases.
Moreover, we observed that both SH15~\cite{shih2015reflection} and AR17~\cite{arvanitopoulos2017single} tend to have a noticeable colour shift and a significant loss of structure; as is visible on (B) the green pole. In particular, we highlight the green pole in (B), in which only our approach was clearly able to maintain the fine details.%  A closer inspection shows that our approach sidesteps the said effects, as is visible in the detail of (d).

We observe that the deep learning based solution FAN17~\cite{fangeneric} shows good edge preservation, but often fails to correctly reproduce colour and texture, and produces noticeable artefacts. This will be discussed further in (E2b).
Overall, out of the evaluated model-based single-image reflection removal techniques, our approach consistently yields the most visually pleasing results.
These observations are confirmed by further examples in Section II of the Supplemental Material.

\begin{figure*}[h!]
\centering
\includegraphics[width=1\textwidth]{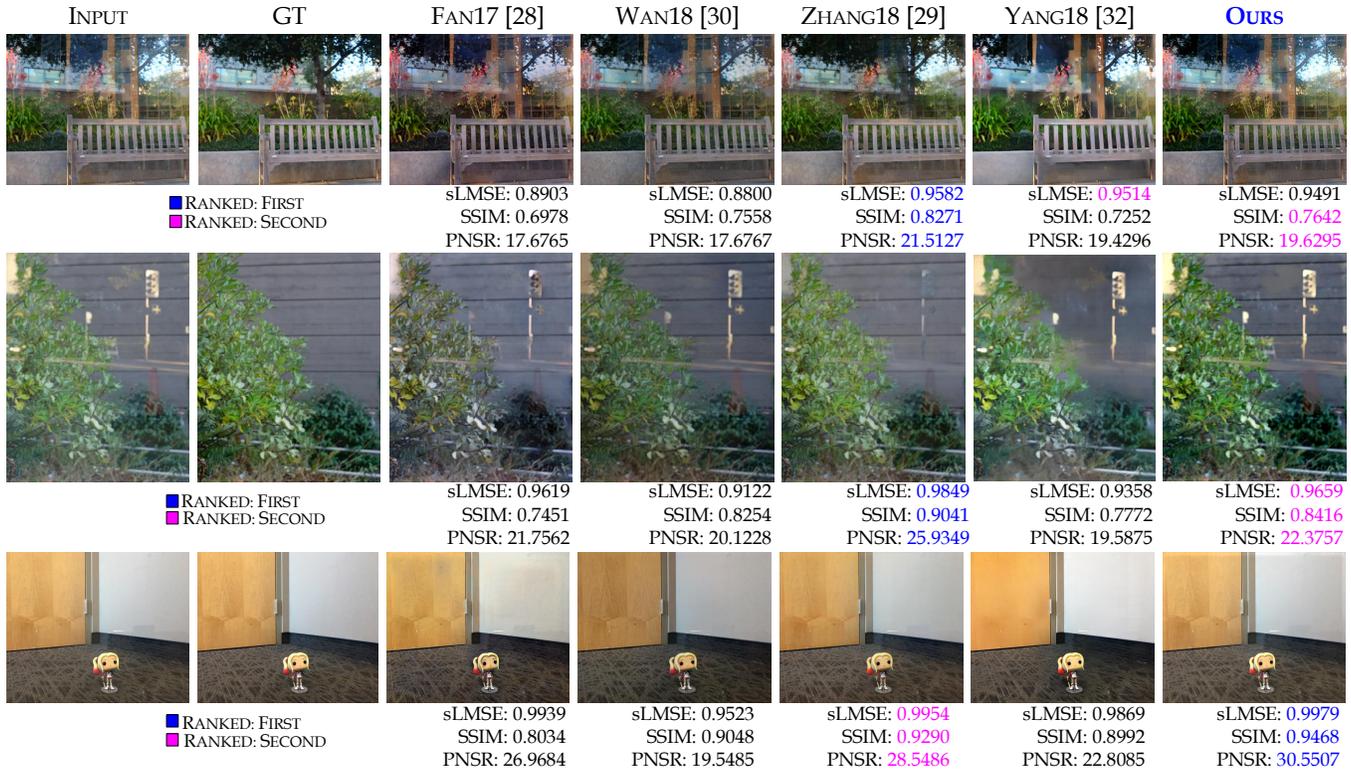}
\caption{(E2b). Visual and numerical comparison of our technique vs. Deep-learning techniques on a selection of images from the Berkley dataset. Details are better appreciated on screen.}
\label{fig::OURSvsDLBerk}
\end{figure*}

\begin{table*}[t!]
\centering
 \resizebox{0.75\textwidth}{!}{
\begin{tabular}{lccccc}
\multicolumn{6}{c}{\cellcolor[HTML]{EFEFEF} \textsc{The Berkley Dataset}}                                                                                                                                                                                                                                                                       \\ \cline{2-6}
\multicolumn{1}{l|}{}                & \multicolumn{1}{l|}{\cellcolor[HTML]{EFEFEF} \textsc{FAN}17~\cite{fangeneric}}   & \multicolumn{1}{l|}{ \cellcolor[HTML]{EFEFEF} WAN18~\cite{wan2018crrn}}  & \multicolumn{1}{l|}{ \cellcolor[HTML]{EFEFEF} ZHANG18~\cite{zhang2018single}}                                                         & \multicolumn{1}{l|}{\cellcolor[HTML]{EFEFEF} YANG18~\cite{yang2018seeing}}                                                & \multicolumn{1}{l|}{{ \cellcolor[HTML]{EFEFEF} \text{   } \textbf{  OURS  } \text{   } }}    \T\B   \\ \hline
\multicolumn{1}{|l|}{\cellcolor[HTML]{EFEFEF} {sLMSE}} & \multicolumn{1}{c|}{0.8407}  & \multicolumn{1}{c|}{0.8090} & \multicolumn{1}{c|}{\cellcolor[HTML]{F6C9F8}\textbf{0.8638}}                         & \multicolumn{1}{c|}{\cellcolor[HTML]{FFFFFF}{\color[HTML]{000000} 0.8398}} & \multicolumn{1}{c|}{\cellcolor[HTML]{96FFFB}\textbf{0.8647}}   \\ \hline
\multicolumn{1}{|l|}{\cellcolor[HTML]{EFEFEF} {SSIM}}  & \multicolumn{1}{c|}{0.7022}  & \multicolumn{1}{c|}{0.6982} & \multicolumn{1}{c|}{\cellcolor[HTML]{96FFFB}\textbf{0.7923}}                         & \multicolumn{1}{c|}{0.6911}                                                & \multicolumn{1}{c|}{\cellcolor[HTML]{F6C9F8}\textbf{0.7315}}    \\ \hline
\multicolumn{1}{|l|}{\cellcolor[HTML]{EFEFEF} {PNSR}}  & \multicolumn{1}{c|}{18.2989} & \multicolumn{1}{c|}{18.300} & \multicolumn{1}{c|}{\cellcolor[HTML]{96FFFB}{\color[HTML]{333333} \textbf{21.6203}}} & \multicolumn{1}{c|}{17.8673}                                               & \multicolumn{1}{c|}{\cellcolor[HTML]{F6C9F8}\textbf{18.7833}}   \\ \hline
\multicolumn{3}{c}{\crule[cyan]{0.3cm}{0.3cm} \textsc{Ranked First}}                                                                  & \multicolumn{3}{c}{\crule[pink]{0.3cm}{0.3cm} \textsc{Ranked Second}}
\end{tabular}
}
\caption{(E2b). Numerical comparison of our technique vs. Deep-learning techniques for the entire Berkley dataset. The numerical values are computed as the averages of the similarity metrics over all images.}
\label{tab:measure-outcome-deep-learning}
\end{table*}

For a more detailed quantitative analysis, we report the global results in Table \ref{tab:measure-outcome}. The displayed numbers are the average of the image metrics across the whole body of `solid-object' files in the dataset, in order to understand the general behaviour and performance of the algorithms.

We observe that both AR17~\cite{arvanitopoulos2017single} and our approach  outperform the remaining algorithms with respect to sLMSE. With respect to SSIM and PNSR, we also achieve significant improvements over most state-of-the-art techniques, most notably over the similar technique AR17~\cite{arvanitopoulos2017single}. The only other approach evaluated here which performs similarly well is the deep learning approach  FAN17~\cite{fangeneric}. As was discussed above, a closer look at single images shows occasional difficulties of this approach, and the more reliable performance of our model-based method.

\begin{figure*}[t!]
\centering
\includegraphics[width=1\textwidth]{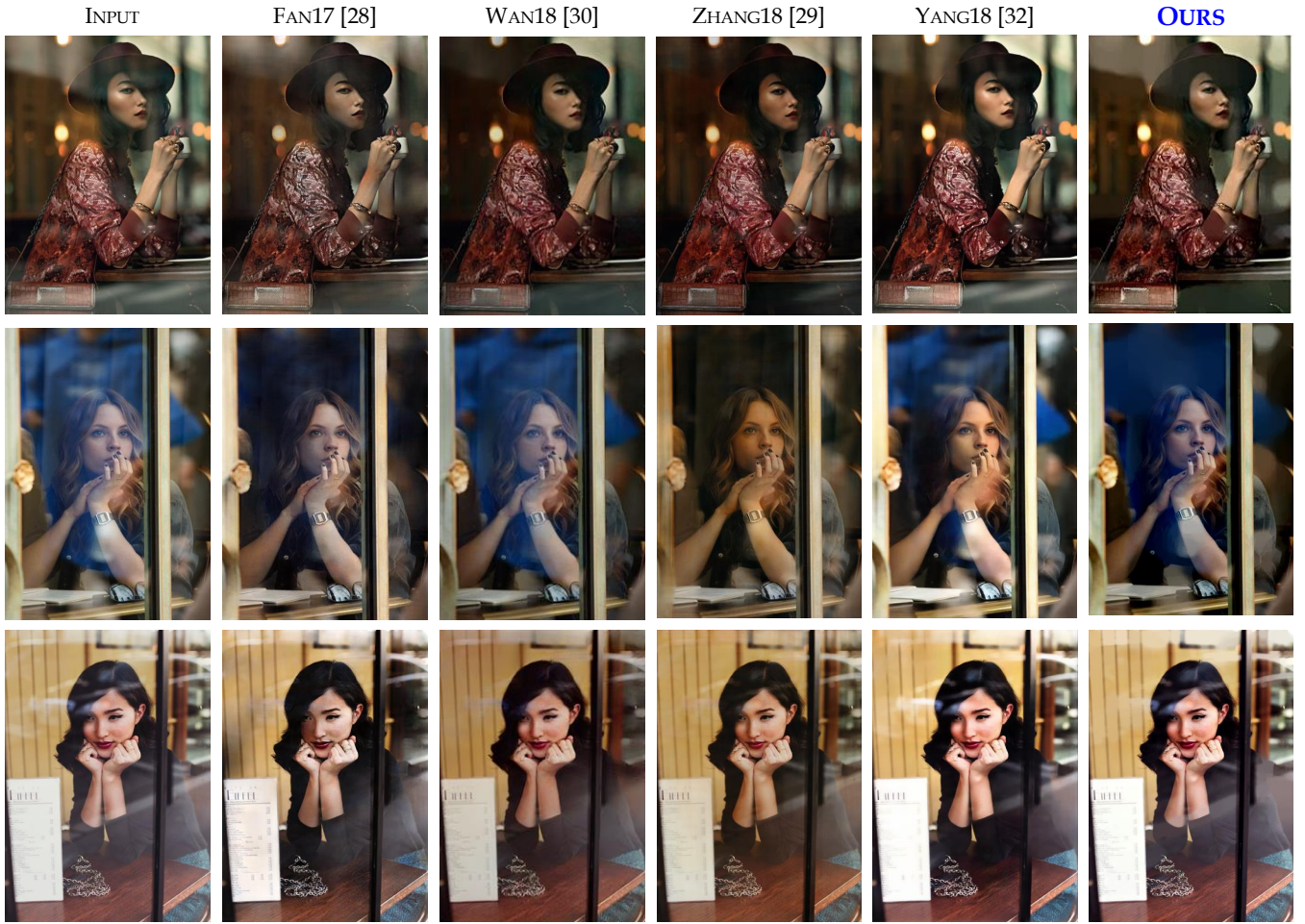}
\caption{(E2b). Comparison of our technique vs Deep-Learning techniques on real-world images. We note that our technique is able to suppress the reflections while avoiding flattening in the skin tone and avoiding false-colour effects. This is an example of our motivation in Observation 2: colour flattening on the skin is much more noticeable than the same effect on the props. Images are from the real-world dataset \cite{fangeneric} and no ground truths are available.}
\label{fig:oursVsDLRW}
\end{figure*}

\medskip
\textbf{(E2b).} Having extensively compared our new method to model-based approaches in (E2a), we now present a detailed comparison against recent advances in single-image reflection removal based on deep-learning. We compare against FAN17~\cite{fangeneric}, WAN18~\cite{wan2018crrn}, ZHANG18~\cite{zhang2018single} and YANG18~\cite{yang2018seeing} on both the Berkeley dataset and real-world images.

Having used FAN17 \cite{fangeneric} as a benchmark for comparison in (E2a), we first present a further comparison of this method against our technique. Indeed, from Table \ref{tab:measure-outcome}, it may appear that FAN17 produces output of a similar quality to our technique. However, we notice that the outputs displayed in Fig. \ref{fig::oursVSstateoftheart} suggest that our method produces visually nicer results; to validate this, we present further experiments in Fig. \ref{fig::OURSvsFAN}. The images displayed are two cases from the SIR$^2$ dataset, in which we observe difficulties similar to those in Fig. \ref{fig::oursVSstateoftheart}.  In Fig.s \ref{fig::oursVSstateoftheart}A, \ref{fig::OURSvsFAN}A, FAN17 has wrongly identified a specular reflection in the transmitted layer as belonging to the reflected layer, producing unpleasant artefacts. We also highlight incomplete reflection removal in the examples in Fig. \ref{fig::OURSvsFAN}, false-colour effects in Fig.s \ref{fig::oursVSstateoftheart} and \ref{fig:OURSvsSIRS}B, and unwanted colour flattening in Fig. \ref{fig::OURSvsFAN}A.

Next, we present a visual comparison of a selection of images from the Berkeley dataset in Fig. \ref{fig::OURSvsDLBerk}. The images include the values of the similarity metrics compared to the ground truth in each case. We observe that FAN17~\cite{fangeneric}, WAN18~\cite{wan2018crrn} suffer from poor colour retention in these test images, while YANG18~\cite{yang2018seeing} induces a significant amount of blurring (see the door in the bottom picture, and the edges of the plant in the middle one). ZHANG18~\cite{zhang2018single} performs very well both visually and numerically, although the quality of its performance somewhat decreases when compared on a different dataset as we do in Fig. \ref{fig:oursVsDLRW}. Our method readily competes with ZHANG18~\cite{zhang2018single} in terms of similarity metrics, but also is able to preserve structure and color much better than the remaining approaches, while still removing a comparable amount of the reflections.

In Table \ref{tab:measure-outcome-deep-learning} we present the similarity measures which are computed as the average over all images in the Berkeley dataset. With respect to sLMSE, our method outperforms all other techniques, in particular FAN17~\cite{fangeneric}, WAN18~\cite{wan2018crrn} and YANG18~\cite{yang2018seeing} by a significant margin. With respect to SSIM and PNSR, our method performs similarly well, and places second behind ZHANG18~\cite{zhang2018single}.

Finally, we test all of the DL methods on a selection of real-world images in Fig.~\ref{fig:oursVsDLRW}. We observe that most of the competing methods suffer from poor colour preservation, which is especially visible in ZHANG18~\cite{zhang2018single} with respect to the skin colour in middle and upper image, and incomplete removal of the reflections. In FAN17~\cite{fangeneric} especially we notice the introduction of artefacts on the arms in the top picture and nearby the head in the bottom one. Our method, while not completely removing the reflections, still ensures good preservation of colour and important structure, and hence in terms of output quality readily competes with the deep-learning based methods. Additional experiments, which further validate this conclusion, may be found in Section III of the Supplemental Material.

The above comparison against the most recent deep learning approaches for this problem, demonstrates that \textit{at this point in time, our model-based method readily competes with deep learning in terms of output quality}. The authors note that traditionally, deep learning has achieved ground breaking success in tasks involving labelling or classification \cite{krizhevsky2012imagenet,lecun2015deep}. The good visual results generated by deep network usually benefit from the statistical information covered in the large body of training samples. However, a  plain fully convolutional neural network does not impose the same kind of rigid and intuitive constraints as model-based approaches; for example, piecewise smoothness is not enforced. Such a limitation in the deep network results in inconsistent reflection removal within a single image, as seen in Fig.s \ref{fig::OURSvsFAN}, \ref{fig::OURSvsDLBerk}, \ref{fig:oursVsDLRW}. While in this paper the deep-learning based techniques provide an important benchmark, their classification as `single-image' techniques raises definitional issues that might be interesting for the community to discuss. This discussion can be found in Section V of the Supplemental Material.

\begin{figure*}[t!]
\centering
\includegraphics[width=1\textwidth]{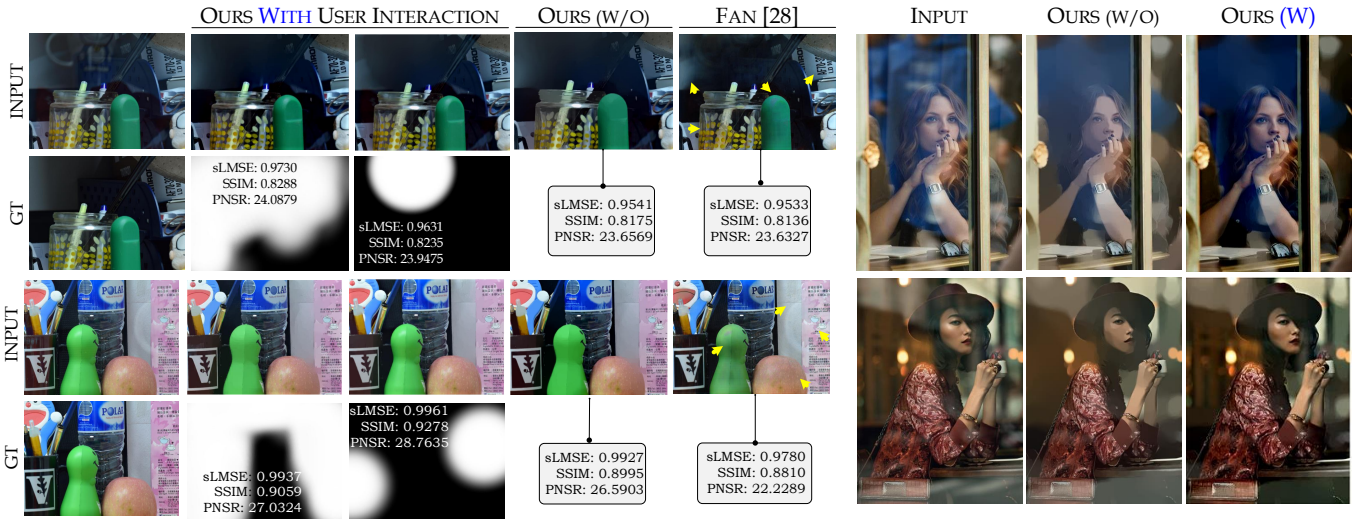}
\caption{(E3). From left to right: The impact of the user-interaction on the outputs computed by OUR approach (with and without user interaction), with FAN\cite{fangeneric} as a benchmark. Examples of cases where region selection leads to noticeable qualitative improvements in avoiding flattening.}
\label{fig:oursVsours}
\end{figure*}

\begin{figure*}[t!]
\centering
\includegraphics[width=1\textwidth]{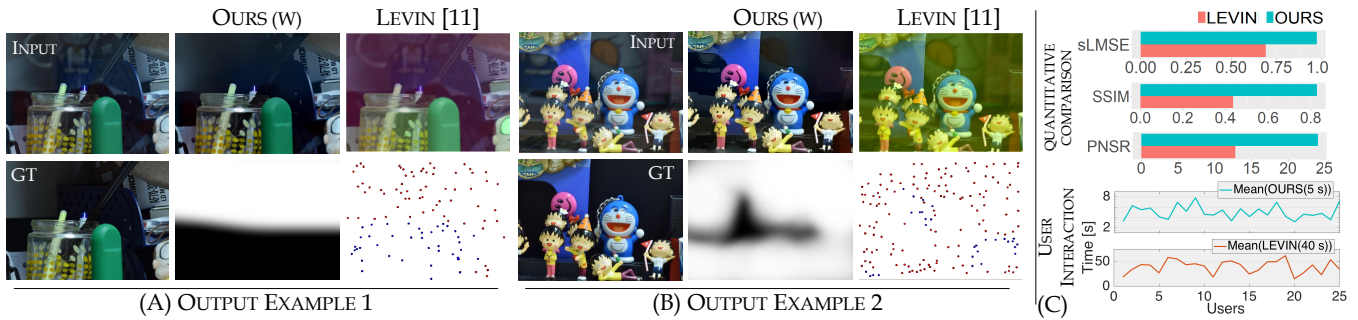}
\caption{(E4). (A-B): Visual comparison of the user-interaction schemes in LEVIN~\cite{levin2007user} and OURS based on a specific example\\(C): Quantitative comparison of the two schemes on the solid object corpus of the SIR$^2$ dataset, and user-interaction time based on a selection of images from this dataset.}
\label{fig:oursVsLevin}
\end{figure*}

\textbf{(E3).} In  Fig.~\ref{fig:oursVsours}, we analyse the impact of the user-interaction, again including FAN17~\cite{fangeneric} as a baseline for comparison. In the first subfigure, we present the results of our approach without region selection, and with both crude and detailed region selection. Without region selection, there is noticeable blurring and flattening: see, for example, the green object in the first example and the apple in the second. However, even with very crude region selection, our technique is able to mitigate these to produce a visually better result. In the second subfigure, we show the result of our technique with and without region selection on two examples from the real-world dataset where region selection makes a substantial visual difference to the output. In both cases, without region selection, the output has a lot of colour flattening on the skin of the model, leading to a very unnatural and unrealistic output. We therefore conclude that \textit{even very crude selection of the reflection regions results in good reflection removal, and that crude region selection noticeably improves on no region selection}.  This justifies our claim of a providing a simple and effective user-interaction scheme.

\textbf{(E4).} We also compare our method to the existent user-interaction by Levin \cite{levin2007user}. We demonstrate that in comparison, our method produces qualitatively and quantitatively better results, while requiring significantly less effort from the end-user. This underlines one of the main messages of this paper, that \textit{we provide a simple user-interaction method, which gives a significant improvement in the quality of the output}.

In Fig.~\ref{fig:oursVsLevin} we compare the amount of user interaction required and the quality of the resulting output for both methods. Firstly, in the bottom half of (A-B), the user-interaction for both methods is shown. For our method, the user is asked to determine the location of reflections in the image by marking the rough location in white; several examples of this user-selection are provided in Section IV of the Supplemental Material. In Levin's approach, the user is asked to select foreground gradients in red and background gradients in blue. We can also see the corresponding output of the algorithm, which can be visually observed to be significantly improved using our method.

In Fig.~\ref{fig:oursVsLevin}~(C) we compare the specific effort of user-interaction between Levin \cite{levin2007user} and our proposed method. For this we asked a group of 25 colleagues to perform the user-interaction on both schemes and try to achieve the best quality removal as quickly as possible. We observe that, on average, our approach took our colleagues around 5 seconds per image, while Levin's method required around  40 seconds, an increase of around $700\%$. The corresponding quantitative results can be seen in the upper half of Fig.~\ref{fig:oursVsLevin}~(C). The numerical values are the metrics averaged over the entire output from 25 users working on the solid-object dataset. In particular each user was given 6 different settings (3 types of focus and 3 types of thickness) of reflections for each of the 20 images in the dataset, and was then asked to perform the user selection for both methods. We see that the similarity metrics are significantly improved using our new method. This shows that our method \textit{requires significantly less effort from the end user than other existent approaches}, while at the same time significantly improving the quality of reflection removal.

\iffalse
\textbf{(E5).}
In Fig. \ref{fig::failureCases}, we present some cases where our technique fails to satisfactorily remove the reflection. In these cases, the strength of the reflection layer is comparable to that of the transmitted layer, corresponding to $w\approx 0.5$ in the model (\ref{eq: model of noisy image}). Moreover, the reflection layer also contains sharp edges, and so the modelling hypotheses are violated. Even in these cases, we observe that our outputs avoid the noticeable colour shifts resulting from the technique of \cite{arvanitopoulos2017single}. The technique of Fan et al. \cite{fangeneric} also performs poorly in these cases: the first and second images are unchanged, and the third introduces artefacts while failing to suppress the reflection.\fi

\section{Conclusions}
This paper addresses the challenging problem of single image reflection removal. We propose a technique in which two novelties are introduced to provide reflection removal of higher quality. The first is an \emph{spatially aware prior term, exploiting low-level user interaction}, which tailors reflection suppression to preserve detail in reflection-free areas. The second is an \emph{$H^2$ fidelity term}, which combines advantages of both $L^2$ and Laplacian fidelity terms, and promotes better reconstruction of faithful and natural colours. Together, these result in better preservation of structure, detail and colour. We demonstrate the potential of our model through quantitative and qualitative analyses, in which it produces better results than all tested model-based approaches and readily competes with recent deep learning techniques.
Future work might include the use of deep learning techniques to automatically select regions, which would avoid the need for user interaction, while preserving many of the advantages of our technique.

% use section* for acknowledgment
\section*{Acknowledgment}

This work was supported by the UK Engineering and Physical Sciences Research Council (EPSRC) grant EP/L016516/1 for the University of Cambridge Centre for Doctoral Training, the Cambridge Centre for Analysis. Support from the CMIH University of Cambridge is greatly acknowledged.

\bibliographystyle{IEEEtran}
\bibliography{bibliographyV2}

\begin{table*}[t!]
\begin{tabular}{c}
\LARGE \textsc{Supplemental Material for the Article: }                                                                                                                                                                                                                                 \\ \\ \\
\begin{tabular}[c]{@{}c@{}} \LARGE Mirror, Mirror, on the Wall, Who's Got the Clearest Image of Them All? \\    \LARGE A Tailored Approach to Single Image Reflection Removal\end{tabular}                                                                                            \\ \\
\begin{tabular}[c]{@{}c@{}} \large Daniel Heydecker$^*$, Georg Maierhofer$^*$, Angelica I. Aviles-Rivero$^*$\\  \large Qingnan Fan, Dongdong Chen, Carola-Bibiane Sch\"onlieb and  Sabine S\"usstrunk\end{tabular}
\end{tabular}
\end{table*}

%\newpage\newpage

%\include{SupplementalMaterial.tex}
%\documentclass[journal,twocolumn]{IEEEtran}
%\documentclass{article}
%\usepackage[total={6.5in, 9in}]{geometry}

%\begin{document}

%\begin{multicols}{2}

%\end{multicols}

%\address{sads}
%\date{}
%\maketitle
%\onehalfspacing
%\spacing{1.5}

%\bigskip
%\noindent

\section{Outline}
This document extends the practicalities and visual results presented in the main paper in order to show further details of our approach and experiments. This is structured as follows.

%This supplemental material extends the practicalities presented in the main paper -this, with the aim of giving further details of our experiments and findings.  The remainder of this document is structured as follows.

\begin{itemize}[noitemsep]
   \item \textbf{Section II:  } We offer further visual results of our  and the state-of-the-art model-based approaches using the SIR$^2$ dataset. As in the main paper, we will that our technique is able to perform noticeably better than other model-based approaches.
  \item \textbf{Section III:} We give further visual comparison of our technique against competing Deep-Learning techniques using the Berkeley dataset\cite{zhang2018single}. This further validates our claim of being able to compete with the DL approaches.
  \item \textbf{Section IV:} We show how our user-interaction technique may be implemented in practice, and demonstrate examples generated by the authors.
  \item \textbf{Section V:} Continuing a point raised in the main text, we discuss a definitional issue over whether competing DL techniques can be considered `single-image'.
  \item \textbf{Section VI:} In the interests of clarity and completeness, we give an explicit definition and motivation of the metrics used for the quantitative
analyses.
\end{itemize}

\bigskip
\section{Supplementary Visual Results with SIR$^2$ dataset} \label{sec::Visualisations}
In this section, we extend the comparison of visual results of Fig. 4 from the main paper. The comparison includes LB14~\cite{li2014single}, SH15~\cite{shih2015reflection}, AR17~\cite{arvanitopoulos2017single}, using FAN17~\cite{fangeneric} as a benchmark.

\begin{figure*}[t!]
\centering
\includegraphics[width=1\textwidth]{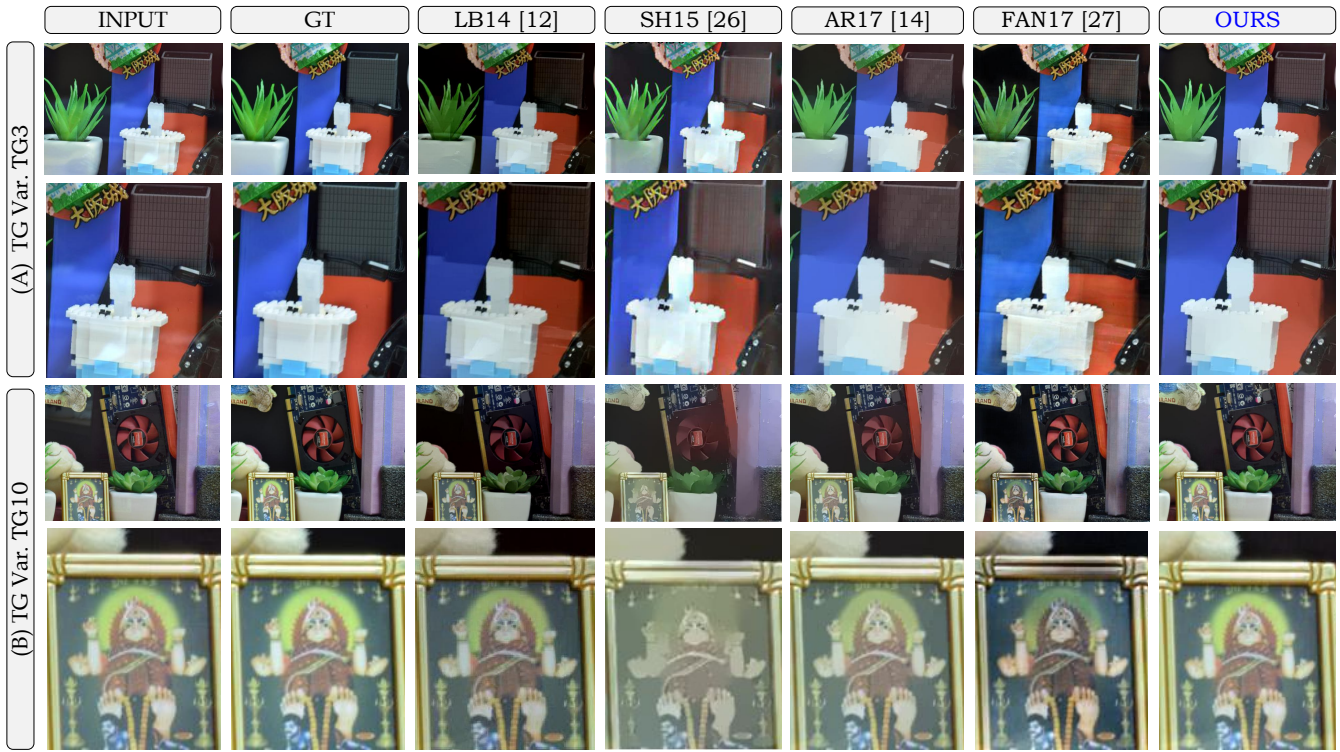}
\caption{Visual comparison against the state-of-the-art of model-based approaches (including FAN17~\cite{fangeneric} as baseline for comparison). The selected frames show variations in shape, colour and texture to appreciate the performance of the compared approaches. Overall, our approach gives a better approximation of $\mathbf{T}$ by preserving colour and structure quality while keeping fine details. Details best appreciated on screen. }
\label{fig:model-based-comparison}
\end{figure*}

\begin{figure*}[t!]

\centering
\includegraphics[width=1\textwidth]{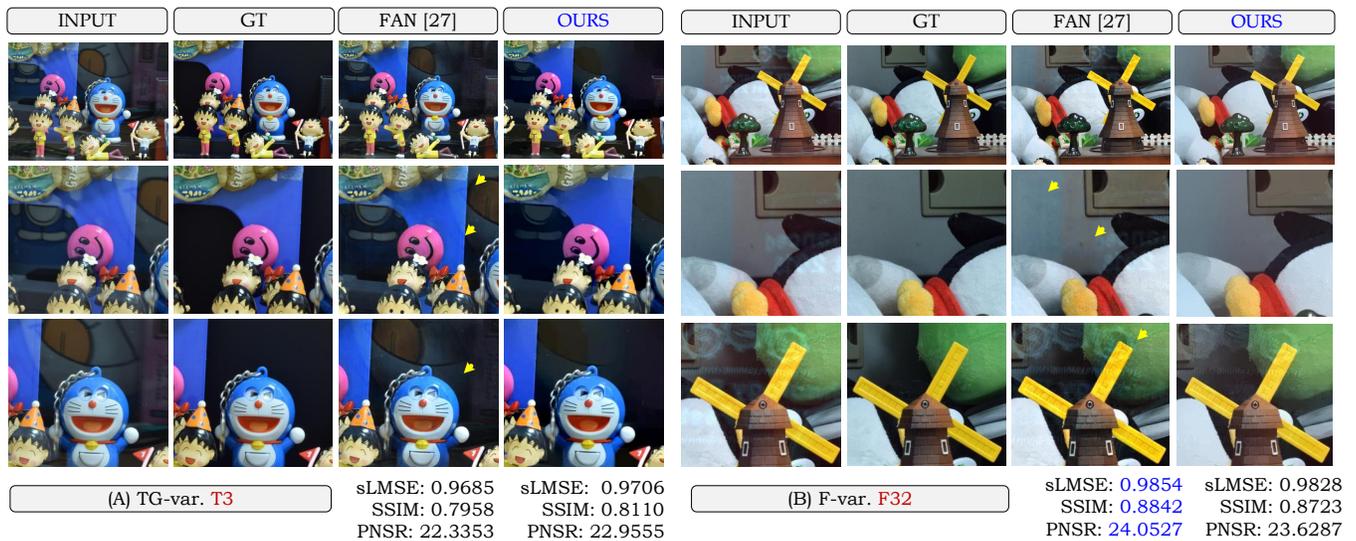}
\caption{Two interesting cases in which we visually and numerically compare our approach against the work of Fan et al. \cite{fangeneric}. We emphasise that even in cases when the metrics are higher for FAN17~\cite{fangeneric}, the output from our algorithm appears visually more appealing and natural. Details best appreciated on screen.}
\label{fig:detail_fan_comparison}
\end{figure*}

In Fig. \ref{fig:model-based-comparison} shows four further examples from the solid object part of the SIR$^2$ dataset and we note that amongst these methods, ours presents the most visually appealing results. In particular we note that LB14~\cite{li2014single} suffers from colour shift at the fan in the third image, and SH15~\cite{shih2015reflection} and AR17~\cite{arvanitopoulos2017single} suffer significant loss of structure in the third and forth images, downsides that are not observed in our technique. Fan on the other hand removes a significantly smaller portion of the reflections in these images -- and this incomplete removal can also be noticed in the further detailed comparison of FAN~\cite{fangeneric} and ours in Fig. \ref{fig:detail_fan_comparison}.

\bigskip
\section{Further Visual Results of Our and DL-based Approaches}

\begin{figure*}[t!]
\centering
\includegraphics[width=1\textwidth]{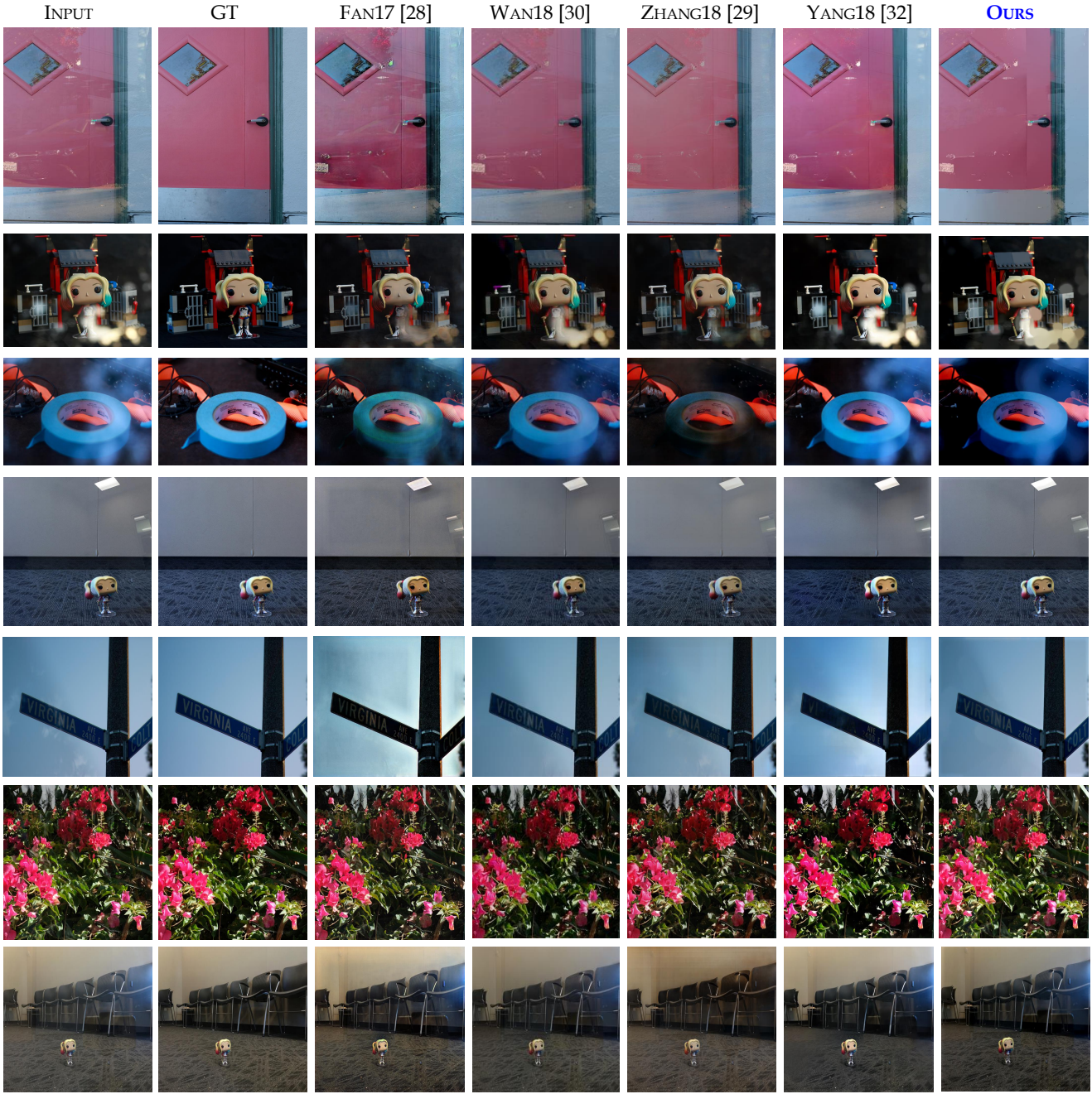}
\caption{Further visual comparison of our technique vs deep-learning techniques, on images from the Berkely dataset \cite{zhang2018single}. Details best appreciated on screen.}
\label{fig: SMoursvsDL}
\end{figure*}

In addition to the experimental results displayed in the main paper, we present some further examples of our technique vs competing DL techniques on elements of the Berkeley dataset \cite{zhang2018single}, displayed in Fig. \ref{fig: SMoursvsDL}. \medskip \\ We see that many of the output images for the competing DL approaches suffer from the same problems described in the main text. The results of FAN17 \cite{fangeneric} introduce very visible artefacts in images 1,4 and 5, which often make the reflections \emph{more} visible than in the input image, and have false-colour effects in images 3 and 7. YANG18 \cite{yang2018seeing} has substantial blurring and loss of detail, which is visible on the carpet in images 4 and 7 and the sign in image 5, and WAN18 \cite{wan2018crrn} introduces substantial blurring throughout. As in the main text, ZHANG18 \cite{zhang2018single} usually performs extremely well on this dataset, but the outputs of images 3 and 7 display noticeable and unpleasant false colour effects. By contrast, our output is able to mitigate the loss of detail and false-colour effects, while competing with the deep-learning approaches in suppressing the reflection layer.

\section{ Supplementary Visual Results of the User Interaction }

\begin{figure*}[t!]
\centering
\includegraphics[width=1\textwidth]{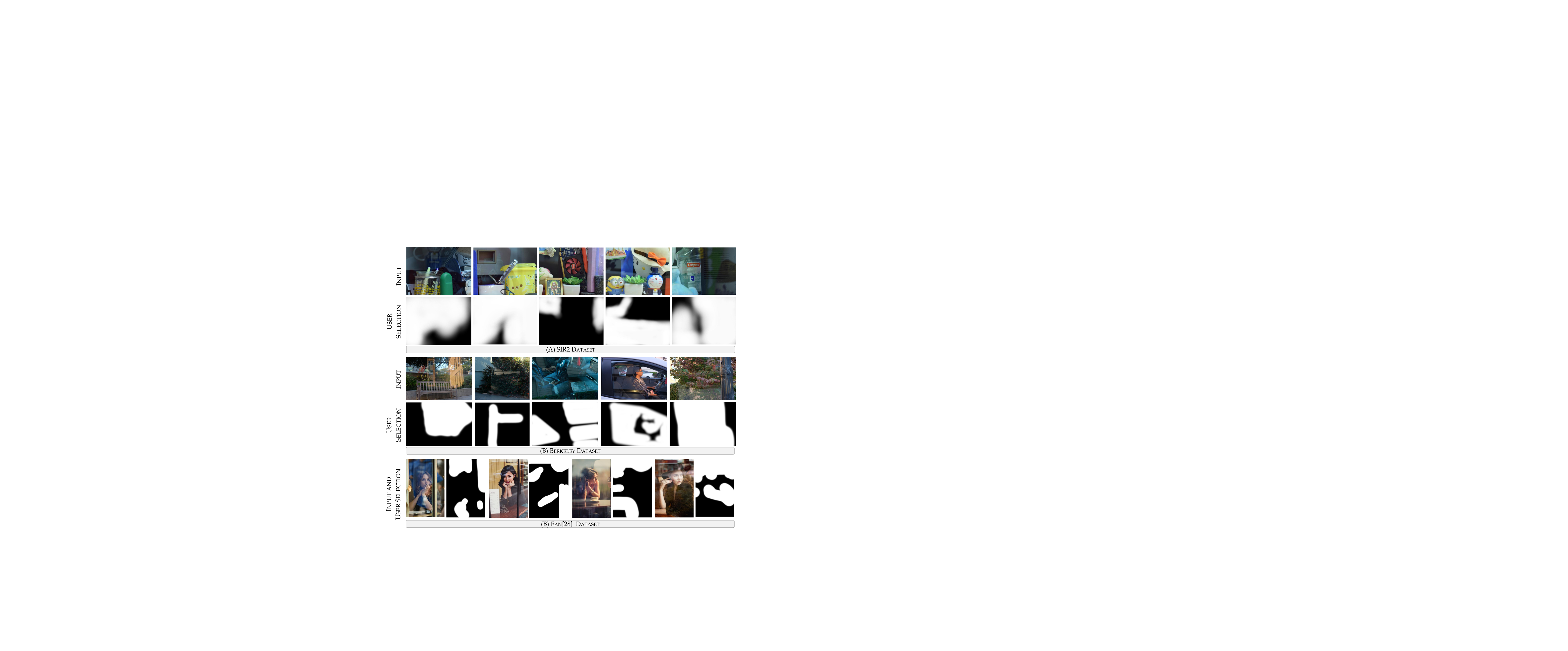}
\caption{Demonstration of the region selection performed by users on a number of images from the three relevant datasets. We show the input and corresponding user selection, these selections where also used in the main experiments of the paper.}
\label{fig:user_interaction}
\end{figure*}

In Fig. \ref{fig:user_interaction} we display a number of examples of the user selection in practise. In particular we display a range of images from all our datasets and show the user-selection as performed in the experiments. In the graph the input images are shown together with the corresponding region selection: Here the user selects a region to be white, if a reflection is seen in that part of the image and black otherwise. This information is then translated (as the relative gray values) into the region selection function $\phi$ as described in section II.A of the main paper.

\section{Is it `Single Image'?}

The need in Deep-Learning for a large set of training data raises the definitional issue of whether the technique could be considered as a `single-image' technique. The definitions of what constitutes a single-image technique reads:

\begin{definition}
 A Single-Image technique is one which uses the information from a single input $\mathbf{Y}$ to extract the transmission layer $\mathbf{T}$.
\end{definition}

In practical terms, this leads to disadvantages similar to those of multiple image techniques; as from a mathematical point of view, DL has the same goal than the multiple-image case: to reduce the strongly ill-possedness problem created in the single-image case.

\section{Definitions of the Metrics }

For clarification purposes, we explicitly define the specific form of the three metrics used in our comparison study.  It is particularly interesting since the results can differ from the ones reported in \cite{wan2017benchmarking} due to the different forms of the metrics. It is to be noted that not all used metrics are explicitly defined in \cite{wan2017benchmarking}, leading to some ambiguity concerning the specific form of, for example, LMSE that is used. Our metrics are computed as follows:

\begin{itemize}
\item sLMSE (Inverted Localised Mean Squared Error) is computed as follows: Let $S$ be an approximation of some ground truth $\hat{S}$. We compute the LMSE as the MSE over patches $S_\omega$ of size $20\times20$, shifted by $10$ each stage, such that
\begin{align}
LMSE(S,\hat{S})=\sum_{\omega}\|S_\omega-\hat{S}_\omega\|_2^2
\end{align}
Then we normalise and produce an inverted measure such that the error measure is 1 if the approximation is good, and zero otherwise.
\begin{align}
sLMSE(S,\hat{S})=\frac{LMSE(S,\hat{S})}{LMSE(S,0)}.
\end{align}

\item SSIM (Structural Similarity Index) is computed in the standard way. Let again $S$ be an approximation of some ground truth $\hat{S}$, and let  $\mu_S$, $\mu_{\hat{S}}$, $\sigma_S$, $\sigma_{\hat{S}}$, $\sigma_{S\hat{S}}$ be the averages, variances and covariance of $S$ and $\hat{S}$ respectively. Then the SSIM is calculated as
\begin{align}
SSIM(S,\hat{S})=\frac{(2\mu_S\mu_{\hat{S}}+c_1)(2\sigma_{S\hat{S}}+c_2)}{(\mu_S^2+\mu_{\hat{S}}^2+c_1)(\sigma_S^2+\sigma_{\hat{S}}^2+c_2)}.
\end{align}
Here, $c_1,c_2$ are variables to stabilise the division in case of weak denominator. In our implementation, these are chosen to be:
\begin{align}
c_1=(0.01*L)^2\\
c_2=(0.03*L)^2
\end{align}
with $L$ being a dynamic range variable that depends on the class of the image (e.g. $L=1$ for type single images).
\item PSNR (Peak Signal-to-Noise Ratio) is also computed in the standard way. Let again $S$ be an approximation of some ground truth $\hat{S}$. Firstly, the full MSE is computed via
\begin{align}
MSE(S,\hat{S})=\frac{1}{N}\|S_\omega-\hat{S}_\omega\|_2^2
\end{align}
where $N$ is the number of total pixels in $S$. We then compute the PSNR as follows:
\begin{align}
PSNR(S,\hat{S})=10\log_{10}\left(\frac{MAX_I^2}{MSE(S,\hat{S})}\right),
\end{align}
where $MAX_I$ is the maximal possible pixel value in the images $S,\hat{S}$ (e.g. 255 for 8-bit images).
\end{itemize}

%\bibliographystyle{IEEEtran}
%\bibliography{bibliographyV2.bib}

%\end{document}
% ----------------------------------------------------------------

% that's all folks
\end{document}